\documentclass{article} % For LaTeX2e
\PassOptionsToPackage{hyphens}{url} % let long URLs in the references break after hyphens
\usepackage{iclr2027_conference,times}

\usepackage{amsmath,amsfonts,bm}

\def\eqref#1{equation~\ref{#1}}
\def\1{\bm{1}}

\DeclareMathAlphabet{\mathsfit}{\encodingdefault}{\sfdefault}{m}{sl}
\SetMathAlphabet{\mathsfit}{bold}{\encodingdefault}{\sfdefault}{bx}{n}

\usepackage{hyperref}
\usepackage{url}
\usepackage{booktabs}
\usepackage{xcolor}
\usepackage[normalem]{ulem}
\usepackage{tikz}
\usepackage{graphicx}
\usetikzlibrary{arrows.meta}

\title{Correct Answers, Invalid Traces: \\ What Verifiable Grade-School Math Reveals About \mbox{Chain-of-Thought} Traces}

\newcommand{\authorblock}[3]{\begin{minipage}[t]{\dimexpr0.5\textwidth-2\tabcolsep\relax}\normalfont\textbf{#1} \\ #2 \\ \texttt{#3}\end{minipage}}
\author{\authorblock{Ratish Puduppully}{IT University of Copenhagen}{rapu@itu.dk}
\And \authorblock{Pranabendu Misra}{Chennai Mathematical Institute}{pranabendu@cmi.ac.in}
\AND \authorblock{Paarth Iyer}{Indian Institute of Technology Jammu}{paarthiyer1234@gmail.com}
\And \authorblock{Durgesh Kalwar}{Arizona State University}{dkalwar@asu.edu}
\AND \authorblock{Vardhan Palod}{Arizona State University}{vpalod@asu.edu}
\And \authorblock{Subbarao Kambhampati}{Arizona State University}{rao@asu.edu}
}

\iclrfinalcopy % arXiv preprint: prints the author block; the header reads ``Preprint'' (see the .sty)
\begin{document}

\maketitle

\begin{abstract}
Chain-of-thought traces are widely read as records of how models reach their answers, informing debugging, agent auditing, and claims about reasoning. Testing this interpretation is difficult because natural-language thinking traces are rarely mechanically verifiable. We revisit it in iGSM, a synthetic grade-school mathematics benchmark designed to study thinking traces and used to support claims of learned reasoning and planning. Crucially, iGSM exposes the exact quantities and dependencies that a correct solution should use, allowing generated traces to be checked programmatically step by step and enabling us to test whether correct answers are reliably accompanied by valid traces.
We first evaluate models trained exclusively on valid, minimal traces. Answer correctness and trace validity nearly coincide in distribution but decouple out of distribution: on the hardest instances, $31.6\%$ of correct answers have invalid traces, over half of which pass all syntactic and arithmetic checks but fail semantic dependency checks. We then intervene on trace supervision. Non-minimal training traces induce non-minimal outputs, while re-asking the same problem with a different query reveals computations inherited from the original query, weakening minimality as evidence of selective planning. Shuffling tokens in $10\%$ of training trace sentences preserves near-clean accuracy even out of distribution despite no trace passing verification. Swapped training traces likewise retain high in-distribution accuracy.
We discuss the implications of these findings for chain-of-thought monitoring  and interpretation in the context of AI safety.

\end{abstract}

\section{Introduction}
\label{sec:introduction}

Chain-of-thought (CoT) prompting~\citep{wei2022chain} became popular as a technique for eliciting step-by-step solutions from large language models (LLMs), and substantially improved performance on a range of reasoning tasks. Subsequent work showed that it can be elicited via simple Zero-shot-CoT~\citep{kojima2022large} prompting: ``Let's think step by step''. Soon CoT was incorporated into training as well. STaR~\citep{zelikman2022star} iteratively generates reasoning traces, retains those that lead to the correct final answer, and fine-tunes the model on these successful traces. These results established that model-generated intermediate reasoning could serve both as additional computation at inference time and as a source of training signal.

Modern reasoning models extend this idea to a much larger scale through reinforcement-learning (RL)-based post-training. Models generate extended CoT trajectories, often referred to as `thinking traces' or `reasoning traces', and are rewarded when those trajectories lead to successful outcomes~\citep{deepseekai2025r1,kimi2025k15}. In prominent published training recipes, this reward is primarily \emph{outcome based}, rather than step-level process rewards. DeepSeek-R1-Zero trains with Group Relative Policy Optimization (GRPO)~\citep{shao2024deepseekmath}, using rule-based rewards for answer correctness and output format without directly considering the content of the reasoning traces~\citep{deepseekai2025r1}. Kimi K1.5 likewise evaluates generated reasoning traces largely by whether they lead to a correct final answer~\citep{kimi2025k15}. Such training therefore directly incentivizes intermediate thinking traces that help produce correct answers, but that does not enforce that the emitted trace is itself a valid derivation of that answer. Process-supervision approaches do attempt to close this gap by assigning feedback to intermediate reasoning steps~\citep{lightman2023verify,wang2024mathshepherd}, but they remain unused in these recipes.

At inference time, these models produce natural-language thinking traces that often resemble a coherent sequence of arguments, deductions, plans, and revisions, making it tempting to read them as an account of what the model is doing and why. This becomes especially consequential when reasoning models are embedded in \emph{agents}: models instructed to adopt particular roles or personas, use tools and take actions in an external environment. In multi-agent settings, many such agents may interact through complex and partially autonomous processes to achieve a given goal, with limited human supervision~\citep{li2023camel,wu2024autogen,chen2024agentverse}. There is also a tendency to conduct post-hoc analysis of the agents' thinking traces, with the intention of understanding how these interactions unfolded; see e.g.~\citep{metr2026huggingface,yao2026competition}.  

With growing emphasis on AI safety and oversight, thinking traces have become a key interface for monitoring and interpreting model behavior. Researchers use them to assess explanation faithfulness~\citep{turpin2023language,lanham2023measuring,chen2025reasoning}, detect undesirable strategies such as reward hacking~\citep{baker2025monitoring}, and interpret complex multi-agent behavior~\citep{metr2026huggingface,yao2026competition}. A recent METR investigation, for example, analyzed the raw thinking traces of a large group of interacting agents and reported coordination, recruitment of other agents into risky experiments, and an agent's trace explicitly contemplating sacrificing its own success for the benefit of the collective~\citep{metr2026huggingface}. Popular coverage has framed such coordination as a form of ``collective intelligence''~\citep{lyons2026agentcollective,wong2026panic}.

The above monitoring, interpretation and analysis all rely, explicitly or implicitly, on an important assumption: the emitted thinking trace is \emph{faithful}, or at least reliably correlated, to how the model arrived at its final answer. However, the fact that a thinking trace is useful to the model for producing a correct answer does not, by itself, establish this. It does not even establish that the emitted trace is a \emph{valid derivation} of the final answer. \emph{Understanding when, how and to what extent thinking traces can be reliably interpreted as records of a model's reasoning is the central motivation of our work}.

\paragraph{Our Results.}
Natural-language thinking traces cannot be checked mechanically, so their
validity is rarely tested directly.
We therefore study iGSM~\citep{ye2024physics21}, a synthetic
grade-school mathematics benchmark, in which the quantities and dependencies underlying each problem are known and model generated traces can be checked programmatically to identify the causes behind why a trace is invalid.
This lets us separate two questions that are often conflated:
\emph{Was the final answer correct?} and \emph{Was the thinking trace a
valid derivation of that answer?}

Prior work interprets the out-of-distribution generalization, minimal
solutions, and hidden-state probes of models trained on iGSM as evidence of
learned reasoning and planning~\citep{ye2024physics21}.
We ask what an external observer can justifiably infer about the model’s capabilities and computation from the thinking traces alone.
We therefore ask a precise and directly measurable question: \emph{When a model produces the correct final answer, does the emitted thinking trace constitute a valid derivation of that answer?} If a correct answer is frequently accompanied by an invalid trace, then properties of that trace cannot be used straightforwardly as evidence for the model’s underlying reasoning process.

We carry out a series of experiments to investigate the above and other related questions on the interpretability of thinking traces. We use the setting of \citeauthor{ye2024physics21}: training GPT-2~\citep{radford2019language} size models trained on iGSM problem instances.
Our first experiment evaluates models trained only on standard valid, minimal traces, which we call the clean model. Answer correctness and trace validity nearly coincide in distribution, but diverge as reasoning depth moves out of distribution. On the hardest problems, $31.6\%$ of correct answers are accompanied by invalid traces. Thus, even under the unusually favorable conditions of synthetic data, a fixed trace grammar, clean supervision, and a formally checkable task, \emph{a correct answer does not reliably certify the validity of the trace.}
Further, these failures need not be obvious from a surface inspection: more than half of these invalid traces pass all syntactic and arithmetic checks, and only fail semantic checks based on the problem's dependencies.

The analysis of the clean model also identifies a concern for \emph{interpreting thinking traces under distribution shift}. As problem difficulty increases beyond what was seen during training, the gap between answer correctness and trace validity grows wider.
These failures arise despite training on valid thinking traces, and without any incentives to conceal behavior. 
This could have implications for AI safety, trace monitoring and interpretation: thinking trace reliability on familiar tasks could deteriorate when the input distribution shifts.

Next, we intervene on the trace supervision and on the query. Training on valid but non-minimal traces substantially changes the minimality of generated
solutions, while re-asking the same problem about a different quantity reveals
computations inherited from the original query. These results show that
minimality alone is insufficient evidence that the model independently selected
exactly the computations required by the current question.

Finally we consider stronger interventions that destroy trace semantics more directly.
Models trained on traces swapped between unrelated problems still answer $82\%$ of in-distribution problems correctly, despite never emitting a valid trace. Next, models trained on traces where $10\%$ tokens of the prefix sentences are randomly shuffled, nearly preserve clean-traces model accuracy, even out of distribution, again with no generated trace passing verification. Larger shuffling progressively harms generalization. 
The above results along with the results on non-minimal traces establish two points. First, a trace can help the model while being far from a valid derivation. Second, features of the trace that look like reasoning, such as minimality, and coherence can be inherited from the training data rather than worked out by the model.

Our conclusions are deliberately narrower than ``chain-of-thought does not matter.'' 
An invalid emitted trace does not show that no useful reasoning occurred internally, and our experiments do not measure whether the trace is a faithful record of the computation that produced the answer. Rather, they show that answer correctness, trace validity, and other properties such as minimality can come apart, and that their
relationship deteriorates under distribution shift. This is important when the thinking traces are used as evidence about how a model reasons or as a signal for interpretation and monitoring. Recent work has proposed monitoring reasoning traces for reward hacking and other undesirable
behavior~\citep{baker2025monitoring,korbak2025monitorability}, while other
studies find that important causal influences may go unmentioned in the
generated reasoning~\citep{chen2025reasoning}. A setting such as iGSM, in which
trace validity can be checked exactly, therefore provides a controlled
test environment of a condition that any stronger claim about the meaning of
thinking traces must satisfy.

\section{Related Work}
\label{sec:related}
Grade-school mathematics, and GSM8K in
particular~\citep{cobbe2021training}, is a foundational benchmark for
reasoning in language models, and step-by-step solutions there play a dual
role: they improve answer accuracy and are read as visible evidence of the
reasoning that produced the answer. \citet{ye2024physics21} introduced iGSM,
a synthetic counterpart whose every problem carries its full dependency
structure, and interpreted out-of-distribution generalization, minimal
solutions and hidden-state probes of models trained on it as evidence that
the models learn to reason and plan. Later work used iGSM to study error
correction~\citep{ye2024physics22} and extended the graph-based construction
to larger problems~\citep{zhou2025gsminf}.

Several findings caution against reading traces as reliable accounts of a
model's reasoning. Plausible CoT explanations can omit factors that causally
influence the answer~\citep{turpin2023language,lanham2023measuring,chen2025reasoning}.
More recently, models have produced correct answers while emitting
semantically invalid traces in maze planning~\citep{valmeekam2026beyond} and
in question answering~\citep{bhambri2026interpretable}. Those studies used
benchmarks built for the purpose. iGSM, in contrast, is an existing
benchmark designed and used in an influential prior effort that arrived at
largely positive conclusions about trace semantics, which lets us test those
conclusions on their own evidence. In chain-of-thought monitoring for AI
safety~\citep{baker2025monitoring,korbak2025monitorability}, an outside monitor reads the trace and looks for problems, which is the role our checker plays. This works only if the trace tells the monitor enough about how the model got to its answer. Our results show how often the trace is invalid even when the answer is right. Appendix~\ref{sec:further-related} extends
this discussion.

\section{Background}
\label{sec:background}

GSM8K~\citep{cobbe2021training}
is a widely used benchmark for evaluating the mathematical reasoning abilities of language models. It 
contains $8.5$K grade-school mathematical word problems with natural-language solutions.
Although the arithmetic involved is elementary, solving these problems often
requires identifying relevant quantities, understanding their dependencies, and
performing a sequence of computations.
\cite{ye2024physics21} introduced \emph{iGSM}, a 
GSM8K-like synthetic dataset, to study how language models learn such multi-step reasoning. Synthetic generation is useful for several reasons. First, the
training and evaluation distributions are completely controlled, substantially
reducing concerns about data contamination and memorization of existing
problems. Second, the generator can produce a very large and diverse collection
of problems; \citeauthor{ye2024physics21} estimate more than $90$ trillion distinct solution
templates even after ignoring numerical values, surface forms, variable names,
sentence ordering, and irrelevant parameters. Third, properties such as
reasoning length can be explicitly controlled, allowing models to be trained on
shorter problems and evaluated on longer ones. Finally, the generator retains
the complete symbolic computation underlying every problem, and hence can
produce a valid reference trace and automatically
check generated reasoning traces.

\begin{figure}[t]
\centering
\begin{tikzpicture}[scale=0.95, every node/.style={transform shape, font=\footnotesize},
  ent/.style={draw, rounded corners, inner sep=3pt, minimum height=1.5em},
  par/.style={draw, rounded corners, inner sep=3pt, minimum height=1.5em, align=center},
  abs/.style={par, fill=gray!15},
  unn/.style={par, densely dashed},
  rng/.style={draw, circle, inner sep=1.5pt},
  lab/.style={font=\scriptsize, fill=white, inner sep=1pt},
  >=Latex, thick]
\begin{scope}
  \node[font=\bfseries] at (2.0,3.3) {(a)};
  \node[anchor=east, gray] at (-0.4,2.4) {habitat};
  \node[anchor=east, gray] at (-0.4,1.2) {animal};
  \node[anchor=east, gray] at (-0.4,0.0) {organ};
  \node[ent] (salt) at (1.0,2.4) {Salt Marsh};
  \node[ent] (est)  at (3.7,2.4) {Estuary};
  \node[ent] (man)  at (1.0,1.2) {Manatee};
  \node[ent] (wal)  at (3.7,1.2) {Walrus};
  \node[ent] (ile)  at (0.1,0.0) {Ileum};
  \node[ent] (sal)  at (2.4,0.0) {Salivary Glands};
  \draw[->] (salt) -- (man);
  \draw[->, densely dashed] (est) -- (wal);
  \draw[->] (man) -- (ile);
  \draw[->] (man) -- (sal);
\end{scope}
\begin{scope}[xshift=6.6cm]
  \node[font=\bfseries] at (3.0,3.3) {(b)};
  \node[rng] (rng) at (0.0,2.4) {RNG};
  \node[par] (w)   at (0.0, 1.0) {Manatee's\\Ileum};
  \node[unn] (u)   at (0.0,0.0) {Estuary's\\Walrus};
  \node[par] (A)   at (2.7,2.4) {Salt Marsh's\\Manatee};
  \node[par] (Q)   at (2.7,0.0) {Manatee's\\Salivary Glands};
  \node[abs] (z)   at (5.5,2.4) {Salt Marsh's\\Organs};
  \node[abs] (m)   at (5.5,0.6) {Manatee's\\Organs};
  \draw[->] (rng) -- node[lab]{\footnotesize 18} (w);
  \draw[->] (rng) -- node[lab]{\footnotesize +3} (A);
  \draw[->, densely dashed] (rng.west) .. controls (-1.3,2.4) and (-1.3,0.0) .. node[lab]{\footnotesize 2} (u.west);
  \draw[->] (w) -- (A);
  \draw[->] (A) -- (Q);
  \draw[->] (w) -- (m);
  \draw[->] (Q) -- (m);
  \draw[->] (A) -- (z);
  \draw[->] (m) -- (z);
\end{scope}
\end{tikzpicture}
\caption{The structure graph (a) and dependency graph (b) of the problem in Figure~\ref{fig:example}. Dashed: the stated parameter the query does not need. Shaded: abstract parameters. $\mathsf{RNG}$ supplies the constants.}
\label{fig:example-graphs}
\end{figure}
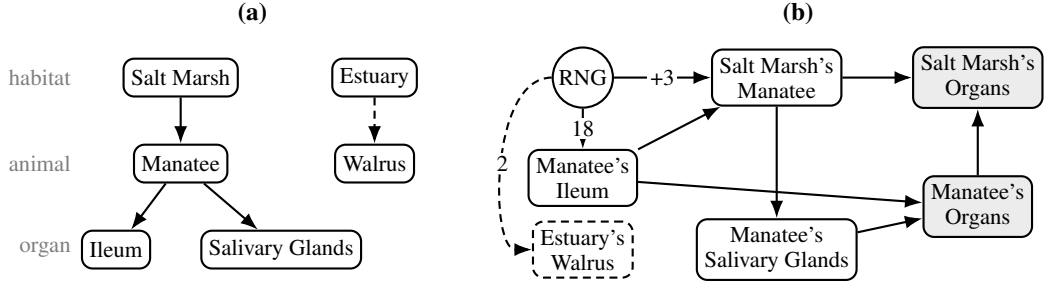

\begin{figure}[t]
\begin{quote}\small
\textbf{Problem.} The number of each Manatee's Salivary Glands equals each Salt
Marsh's Manatee. The number of each Manatee's Ileum equals 18. The number of
each Estuary's Walrus equals 2. The number of each Salt Marsh's Manatee equals
3 more than each Manatee's Ileum. How many Organs does Salt Marsh have?

\textbf{Reference trace.} Define Manatee's Ileum as w; so w = 18. Define Salt
Marsh's Manatee as A; so A = 3 + w = 3 + 18 = 21. Define Manatee's Salivary
Glands as Q; so Q = A = 21. Define Manatee's Organs as m; so m = Q + w = 21 + 18
= 16. Define Salt Marsh's Organs as z; so z = A * m = 21 * 16 = 14.

\textbf{Answer.} 14.
\end{quote}
\caption{A running example: An iGSM problem with its reference trace and answer, as produced by the generator. The sentence about the Estuary's Walrus is not needed for the query; Manatee's Organs and Salt Marsh's Organs are totals implied by the hierarchy and are not stated.}
\label{fig:example}
\end{figure}

\paragraph{Problems, Solutions and Dependency Structures in iGSM.}
Figure~\ref{fig:example} shows an iGSM problem with its reference trace and
answer, which we use as a running example. The problem is a set of sentences, each stating one quantity, in random order,
followed by a query. The trace defines one quantity per sentence, gives it a
one-letter name, and computes it from quantities defined earlier. All
arithmetic is modulo 23, so $21+18=16$ and $21\times16=14$. Two features of
the example matter for what follows. The sentence about the Estuary's Walrus is
never used, and two of the five quantities in the trace, Manatee's Organs and
Salt Marsh's Organs, are not stated in the problem at all.

The \emph{structure graph} $G_{\mathrm{s}}$ is a layered graph whose vertices
are entity categories and whose edges connect entities in adjacent layers. In
the example there are three layers, habitats (Salt Marsh, Estuary), animals
(Manatee, Walrus) and organs (Ileum, Salivary Glands);
Figure~\ref{fig:example-graphs}(a) shows them. Each edge is an \emph{instance
parameter}, the number of the lower entity in each upper entity. ``Salt
Marsh's Manatee'' is the number of manatees in each salt marsh, and
``Manatee's Ileum'' the number of ilea in each manatee. Every sentence of the
problem states one instance parameter, either as a constant or as a function
of other parameters, so the problem text fixes the edges of $G_{\mathrm{s}}$.

The hierarchy also defines quantities that no sentence states. ``Manatee's
Organs'' is the number of organs of any kind in each manatee, the sum over the
organ types the manatee has, here Ileum plus Salivary Glands. ``Salt Marsh's
Organs'' is the number of organs in each salt marsh, the number of manatees per
marsh times the organs per manatee; with several animal types in the marsh the
products would be summed. These are \emph{abstract parameters}. They are never stated;
their definitions follow from the hierarchy, and computing them across layers
involves products as well as sums.

The \emph{dependency graph} $G_{\mathrm{d}}$ is a directed acyclic graph over
the parameters that records which quantities are needed to compute each one
(Figure~\ref{fig:example-graphs}(b)). An edge $a\rightarrow b$ means that $b$
depends on $a$. Constants come from a special vertex $\mathsf{RNG}$, which
supplies both the values of parameters that are simply given (Manatee's Ileum
equals 18) and the constants in equations (the 3 in ``3 more than'').
In the example, the abstract parameter Manatee's Organs depends on the
instance parameters Manatee's Ileum and Manatee's Salivary Glands, and the
abstract parameter Salt Marsh's Organs depends on the instance parameter Salt
Marsh's Manatee and the abstract parameter Manatee's Organs.

Not every stated parameter is needed for the query. Let
$G_{\mathrm{d}}^{\mathrm{nece}}$ be the subgraph of parameters on which the
queried quantity depends. In the example it has five parameters; Estuary's
Walrus is stated but lies outside it (dashed in Figure~\ref{fig:example-graphs}). The reference
trace computes exactly the parameters of $G_{\mathrm{d}}^{\mathrm{nece}}$, in a
topological order, that is, each parameter after all parameters it depends on.
The order need not be unique. In this case, though, Manatee's Ileum must come first and Salt Marsh's
Organs last. Solving a problem thus requires identifying
$G_{\mathrm{d}}^{\mathrm{nece}}$ among the stated sentences and finding a
valid order.

\paragraph{Difficulty.}
iGSM controls problem difficulty through $\mathsf{op}$, the number of
operations in the necessary computation, together with $\mathsf{ip}$, the
number of instance parameters. For a parameter $a$ with in-degree $t$ in
$G_{\mathrm{d}}$, ~\citeauthor{ye2024physics21} define
$\mathsf{op}(a)=\max\{1,t-1\}$, and $\mathsf{op}$ is the sum of
$\mathsf{op}(a)$ over the parameters of $G_{\mathrm{d}}^{\mathrm{nece}}$ other
than $\mathsf{RNG}$. Constant assignments and copies therefore count as one
operation each, as does each binary arithmetic operation. In the example every
parameter has in-degree at most two, so each contributes one operation and
$\mathsf{op}=5$. The generator can filter problems to a specified value of
$\mathsf{op}$. The \textsf{iGSM-med} setting uses $\mathsf{ip}\leq20$ and
trains on $\mathsf{op}\leq15$, with out-of-distribution tests at
$\mathsf{op}\in\{20,21,22,23\}$; the \textsf{iGSM-hard} setting uses
$\mathsf{ip}\leq28$, trains on $\mathsf{op}\leq21$, and tests up to
$\mathsf{op}=32$~\citep{ye2024physics21,ye2024physics22}. All arithmetic is
modulo 23, so every quantity is an integer in $\{0,\dots,22\}$.

\paragraph{Checking a trace.}
Because every quantity has a name and a definition, a generated trace can be
checked against the problem step by step. The evaluator parses the trace, then verifies that every name used was
defined earlier, that each definition uses exactly the inputs the
dependency graph prescribes, that the arithmetic is right modulo 23, and
that the trace ends by defining the queried parameter. A trace passing all
four checks is \emph{valid}; the answer the model states is compared with
the true value separately. Writing $21\times16=15$ would fail the
arithmetic check. Defining Manatee's Organs as $A+w$ instead of $Q+w$
would fail the dependency check even though $A$ and $Q$ share the value
21, because the check compares the parameters a step reads with those
prescribed, not their values. The check does not consider the operation that combines two quantities; this can result in a valid trace with a wrong answer
(Example~9 in Appendix~\ref{sec:examples}).

\section{Experiments and Results}

\paragraph{Setup.}
We follow the iGSM-med recipe of \citet{ye2024physics21}: 124M-parameter
GPT-2-style models are trained from scratch on on-the-fly problems with
$\mathsf{op}\le15$ and $\mathsf{ip}\le20$. Models differ only in the training
trace paired with each problem. We evaluate greedily on 4,096 held-out
instances per level, including out-of-distribution
$\mathsf{op}\in\{20,\ldots,23\}$. A trace is \emph{valid} when it passes all
checks described in Section~\ref{sec:background}. Because answers are computed
modulo 23 and zeros propagate through multiplication, an always-zero predictor
scores about 15\%, which we use as a useful baseline when interpreting low
accuracies.
Results use a single clean model run; our other clean runs, the comparison with \citet{ye2024physics21}, and seed variability are given in
Appendix~\ref{sec:experiment-details}.

\begin{table}[h]
\centering
\begin{tabular}{l rr rrrr}
\toprule
 & \multicolumn{2}{c}{in-distribution} & \multicolumn{4}{c}{out of distribution} \\
\cmidrule(lr){2-3}\cmidrule(lr){4-7}
training trace & op $\le 15$ & op $=15$ & op $=20$ & op $=21$ & op $=22$ & op $=23$ \\
\midrule
clean               & 100.0 & 98.8 & 83.3 & 77.8 & 67.8 & 58.0 \\
swapped             &  81.7 & 19.9 & 16.5 & 16.9 & 16.8 & 17.4 \\
swapped, op-matched &  81.4 & 19.2 & 16.1 & 16.4 & 16.7 & 17.1 \\
\bottomrule
\end{tabular}
\caption{Accuracies for iGSM with clean, swapped traces, and swapped traces with operations matched. The accuracies for each operation in-distribution is in Appendix Table \ref{tab:arms-swapped-per-op}}
\label{tab:arms-swapped}
\end{table}

\paragraph{Swapped traces.}
The models in Table~\ref{tab:arms-swapped} are trained on traces taken from a different,
independently generated problem, either drawn at random (swapped) or
matched in $\mathsf{op}$ so that the trace has the same number of operations as the
valid one would (op-matched). No emitted trace of either model passes the
check at any level, since the trace names parameters of another problem.
In distribution both models nevertheless answer $82\%$ of problems
correctly. That average is dominated by short problems. Split by op
(Table~\ref{tab:arms-swapped-per-op}), the swapped models answer nearly every
problem of up to four operations and $82$--$96\%$ of those with five to
seven, all while emitting traces that belong to other problems; accuracy
then falls with op, to $30\%$ at eleven and $20\%$ at fifteen. Beyond the training distribution their accuracy falls to
$16$--$20\%$, the level of the zero-answer prior.
\begin{table}[h]
\centering
\begin{tabular}{l rr rrrr}
\toprule
 & \multicolumn{2}{c}{in-distribution} & \multicolumn{4}{c}{out of distribution} \\
\cmidrule(lr){2-3}\cmidrule(lr){4-7}
training trace & op $\le 15$ & op $=15$ & op $=20$ & op $=21$ & op $=22$ & op $=23$ \\
\midrule
clean                & 100.0 & 98.8 & 83.3 & 77.8 & 67.8 & 58.0 \\
10\% shuffled        &  99.9 & 98.5 & 83.9 & 78.5 & 69.6 & 60.1 \\
30\% shuffled        &  98.6 & 74.9 & 43.5 & 37.3 & 31.2 & 27.8 \\
50\% shuffled        &  93.5 & 42.9 & 18.4 & 16.8 & 14.6 & 15.0 \\
75\% shuffled        &  83.1 &  0.6 &  0.0 &  0.0 &  0.0 &  0.0 \\
100\% shuffled       &  66.6 &  4.8 &  1.8 &  1.8 &  1.2 &  1.2 \\
swapped              &  81.7 & 19.9 & 16.5 & 16.9 & 16.8 & 17.4 \\
no trace             &  87.6 & 24.7 & 18.0 & 18.1 & 16.9 & 17.7 \\
\bottomrule
\end{tabular}
\caption{Table showing results for training model with shuffled prefix sentences of the trace. The results are compared with models trained on clean traces, on swapped traces, and on no trace (the model emits only the answer).}
\label{tab:arms-shuffled}
\end{table}

\paragraph{Shuffled prefixes.}
The models in Table~\ref{tab:arms-shuffled} are trained on traces whose first $p\%$ of sentences
are replaced by a single block containing the same tokens in random order; the
remaining suffix is left intact. Thus, names and values remain present, but the
corrupted prefix is unreadable. For the running example in
Figure~\ref{fig:example}, $p=30$ shuffles the first two of five sentences into
a block such as
``\emph{='s Man. 21. w 18; 3 +eum Def Salt w so's = +ine =; Def Ale A so 3
Manine w 18ate I =eate as as Marsh}'',
after which the final three sentences remain unchanged. Because corruption ends
at sentence boundaries, the actual token fractions destroyed are
$21,29,45,61,$ and $100\%$ for $p=10,30,50,75,100$.
The models learn to emit the shuffled block followed by well-formed suffixes, so
no generated trace passes the checker. Nevertheless, accuracy degrades only
gradually with corruption. At $p=10$ the model matches the clean model
throughout. At $p=30$ it still reaches $75\%$ at $\mathsf{op}=15$ and $44\%$
at $\mathsf{op}=20$, while a model trained with no trace at all scores $25\%$
and $18\%$ at those levels, and the swapped model $20\%$ and $17\%$. At
$p=50$ the model beats both baselines only in distribution, and at $p\ge75$
the models mostly fail to terminate out of distribution. Thus substantial
answer-solving ability can survive traces that are never valid, although
increasing corruption eventually destroys the useful computation carried by
the trace.

\begin{figure}[t]
\centering
\includegraphics[width=0.8\linewidth]{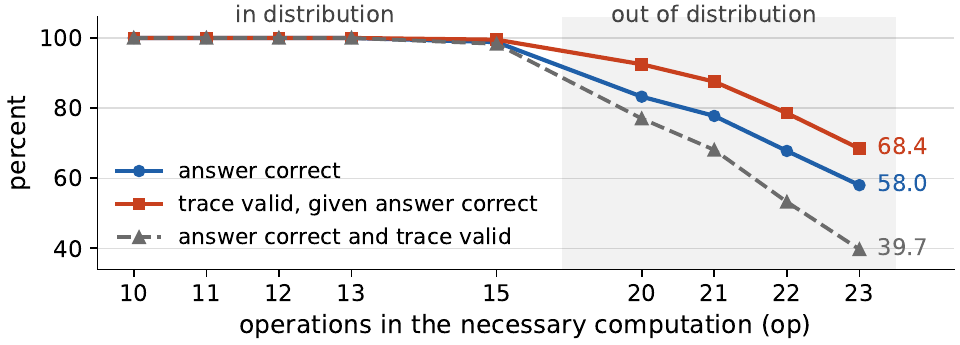}
\caption{Answer accuracy of the clean model at each number of operations, together with the share of instances with both a correct answer and a valid trace, and the share of correct answers whose trace is valid. Inside the training range all three are near 100\%. Beyond it, a growing share of correct answers comes with an invalid trace. Ops 1 to 9, where all three curves are at 100\%, are omitted; Table~\ref{tab:quadrants-clean-full} in the appendix gives the counts.}
\label{fig:pvc-vs-op}
\end{figure}

\begin{table}[t]
\centering
\small
\begin{tabular}{l r r r}
\toprule
 & op $=15$ & op $=20$ & op $=23$ \\
\midrule
correct answers & 4047 & 3410 & 2376 \\
\quad with an invalid trace & 20 & 256 & 750 \\
\quad share of correct answers with invalid traces (\%) & 0.5 & 7.5 & 31.6\\
\midrule
\textbf{Syntactic/arithmetic failures (total)} & 1 & 56 & 366 \\
\quad share of invalid traces (\%) & 5.0 & 21.9 & 48.8 \\
\quad trace does not parse & 1 & 54 & 352 \\
\quad reference, order, or redefinition error & 0 & 2 & 14 \\
\quad arithmetic error & 0 & 0 & 0 \\
\midrule
\textbf{Semantic-only failures (total)} & 19 & 200 & 384 \\
\quad share of invalid traces (\%) & 95.0 & 78.1 & 51.2 \\
\quad defines a name not in the problem & 0 & 3 & 6 \\
\quad a step uses the wrong inputs & 19 & 197 & 377 \\
\quad other dependency mismatch & 0 & 0 & 1 \\
\bottomrule
\end{tabular}
\caption{Invalid traces accompanying correct answers, grouped by the checks they fail. Table~\ref{tab:invalid-correct-causes-full} in the appendix extends it to all five levels, splits the row ``a step uses the wrong inputs'' by how the answer survived, and points to an example of each row.
}
\label{tab:invalid-correct-causes}
\end{table}

\paragraph{Correct answers with invalid traces.}
Figure~\ref{fig:pvc-vs-op} separates answer correctness from trace
validity for the clean model at every op level. It plots answer accuracy,
the share of instances with both a correct answer and a valid trace, and
their ratio, the probability that the trace is valid given a correct
answer. In distribution the first two curves coincide, so
nearly every correct answer comes with a valid trace. They diverge with
reasoning depth: at
$\mathsf{op}=15,20,23$, correct answers with valid traces account for
$98.3\%$, $77.0\%$, and $39.7\%$ of all instances, while invalid traces
accompany $0.5\%$, $7.5\%$, and $31.6\%$ of correct answers, respectively.
Thus a correct answer becomes progressively weaker evidence of a valid
derivation, and doubling the training steps does not change this
(Appendix~\ref{sec:experiment-details}).

Table~\ref{tab:invalid-correct-causes} distinguishes syntactic from semantic failures. A syntactic failure is detectable from the trace alone, because the trace does not parse, uses a symbol it never defined, or gets an equation wrong. A semantic failure passes those checks and shows only when the trace is compared with the problem, because a step reads a quantity the dependency graph does not prescribe or defines a name the problem does not contain. The latter account for $95.0\%$, $78.1\%$, and $51.2\%$ of invalid traces accompanying correct answers at $\mathsf{op}=15,20,23$. Hence, even with exclusively valid training traces, distribution shift leads to outputs that can remain syntactically and arithmetically plausible while no longer constituting valid derivations of the problem. Since clean traces nevertheless substantially improve out-of-distribution accuracy (Tables~\ref{tab:arms-swapped}
and~\ref{tab:arms-shuffled}), computational usefulness and trace validity are
distinct properties. This makes distribution shift an important consideration
for trace-based interpretation and monitoring.

\begin{table}[t]
\centering
\small
\begin{tabular}{l r rr rr}
\toprule
 & & \multicolumn{2}{c}{slack in the problem} & \multicolumn{2}{c}{emitted trace} \\
\cmidrule(lr){3-4}\cmidrule(lr){5-6}
training traces & op & unneeded params & none (\%) & extra params / trace & minimal (\%) \\
\midrule
minimal (clean)      & 15 & 2.66 & 37.8 & $0.00 \pm 0.07$ & 99.9 \\
                     & 20 & 1.84 & 49.9 & $0.01 \pm 0.45$ & 98.7 \\
                     & 23 & 1.35 & 60.0 & $0.13 \pm 2.02$ & 93.8 \\
\midrule
90\% non-minimal     & 15 & 2.66 & 37.8 & $1.20 \pm 2.47$ & 73.9 \\
                     & 20 & 1.84 & 49.9 & $0.99 \pm 2.09$ & 75.2 \\
                     & 23 & 1.35 & 60.0 & $0.77 \pm 1.76$ & 77.7 \\
\midrule
100\% non-minimal    & 15 & 2.66 & 37.8 & $2.45 \pm 2.87$ & 40.2 \\
                     & 20 & 1.84 & 49.9 & $1.79 \pm 2.44$ & 50.8 \\
                     & 23 & 1.35 & 60.0 & $1.35 \pm 2.20$ & 60.9 \\
\bottomrule
\end{tabular}
\caption{Minimality of emitted traces under minimal and non-minimal training supervision. Slack is the number of stated parameters the query does not need (mean over 4,096 problems, and the share of problems with none). We report the unnecessary parameters per trace (mean $\pm$ sd) and the share of traces with no unnecessary parameters.}
\label{tab:minimality}
\end{table}

\begin{table}[t]
\centering
\small
\begin{tabular}{l r l r r r r}
\toprule
world & op & query & accuracy & irrelevant & padded (\%) & unnec.\ params / trace \\
\midrule
standard & 10 & original & 99.9  & 3     & 0.0 & 0.00 \\
standard & 20 & original & 84.5  & 1     & 0.4 & 0.00 \\
standard & 15 & re-asked & 89.6  & 2--9  & 2.6 & 0.14 \\
standard & 20 & re-asked & 86.9  & 2--9  & 3.7 & 0.18 \\
\midrule
wide     & 10 & original & 100.0 & 11    & 0.4 & 0.01 \\
wide     & 20 & original & 67.6  & 6     & 1.3 & 0.01 \\
wide     & 15 & re-asked & 76.6  & 6--15 & 9.7 & 0.70 \\
wide     & 20 & re-asked & 76.3  & 6--15 & 9.6 & 0.77 \\
\bottomrule
\end{tabular}
\caption{Minimality of the clean model's traces in a widened world and for re-asked questions, 1,024 problems per cell; op is that of the problem as generated. Irrelevant sentences are stated sentences the query does not need. Padded traces and unnecessary parameters per trace are counted over correctly answered problems; a padded trace defines at least one unnecessary parameter.}
\label{tab:reask-wide}
\end{table}

\paragraph{Non-minimal training traces.}
\citet{ye2024physics21} interpret iGSM models' shortest solutions as evidence
that they select only parameters needed for the computation, but their training
traces are themselves minimal. To test whether this behavior is independently
inferred, we train models on traces that evaluate every problem parameter in a
random topological order, either for all training examples or for 90\% of them,
with the remaining 10\% minimal. Table~\ref{tab:minimality} shows that
minimality changes sharply with supervision. With 100\% non-minimal training
the model emits 2.5 unnecessary parameters per trace at $\mathsf{op}=15$ and
only 40\% of its traces are minimal, against 99.9\% under standard training;
with 10\% minimal supervision about three quarters are minimal. The minimal
share of the non-minimal model rises with $\mathsf{op}$, to 61\% at
$\mathsf{op}=23$, but only because 60\% of the problems at that level state
no parameter the query does not need; where there is slack, the model
includes nearly all of it. Minimality is thus strongly shaped by the
training-trace distribution rather than fixed by the task.

\paragraph{Is minimality a planning ability?}
In every training problem the question asks for the sink of the dependency
graph, the parameter nothing else depends on (Salt Marsh's Organs in Figure~\ref{fig:example-graphs}(b)), so ``the parameters the
question needs'' and ``the chain that ends at the sink'' are the same set. A
model that simply computes toward the sink therefore looks as if it selects
what the question needs, and the standard evaluation cannot tell the two
apart. In the current setup, it also gives the model little to leave out: at
$\mathsf{op}=20$ half of the problems do not state any parameter which the
query does not need (Table~\ref{tab:minimality}). Two tests can separate them
(Table~\ref{tab:reask-wide}). Widening the world adds sentences to leave
out but keeps the sink. Re-asking moves the
question off the sink while keeping the text.

The first test fixes the number of entities per layer at four and the depth
at four and allows up to 48 stated relations, so that a typical problem
states 6--11 sentences the query does not need, a third of them defining
parameters that depend on the chain. The original-question rows of
Table~\ref{tab:reask-wide} show that minimality survives. With the
original query, the share of correctly answered problems whose trace
defines an unnecessary parameter is 0.4--1.3\%, against 0.0--0.4\% in the
standard setting.

The second test, following \citet{ye2024physics21}, re-points the question
of a generated problem at a randomly chosen parameter, so that the text is
unchanged while the required chain no longer ends at the sink. The re-asked
rows of Table~\ref{tab:reask-wide} show that minimality fails. Among correctly answered problems, re-asking raises the share with a
padded trace from 0.4\% to 3.7\% in the standard setting at op~20, and
from 1.3\% to 9.6\% in the widened world. The extra parameters are parameters of the chain to the sink that the
new question does not need. In the running example, re-asking for Manatee's Salivary Glands
leaves Manatee's Organs and Salt Marsh's Organs unnecessary, and the model
tends to compute them anyway while almost never computing the distractor
Estuary's Walrus. Across all problems, 89--97\% of the unnecessary
parameters the model includes lie on the chain to the sink, though such
parameters are only 18--28\% of those available.

Minimality in iGSM is thus a property of how the questions are posed, and
the standard evaluation cannot tell it from planning because it never
poses a question anywhere else.

\section{Discussion}
\label{sec:discussion}

Our results separate two properties that are easy to conflate: whether a
thinking trace is \emph{useful computation} for producing an answer, and
whether the emitted trace is itself a \emph{correct derivation} of that answer.
Clean traces clearly help with the former. Models trained on clean iGSM traces
generalize far better to problems with increased reasoning depth than models trained on swapped or
heavily shuffled traces (Tables~\ref{tab:arms-swapped}
and~\ref{tab:arms-shuffled}). The swapped and shuffled interventions carry different claims.
The shuffled models show that a trace can be useful while never valid. With
$30\%$ of sentences shuffled the model scores $75\%$ at $\mathsf{op}=15$
against $25\%$ with no trace at all. The swapped models show that a trace
can be pure ritual. The model answers $82\%$ of in-distribution problems
while every trace it writes belongs to another problem. Yet the clean model's advantage does not preserve trace
validity. The share of correct answers with an invalid trace rises from
$0.5\%$ at $\mathsf{op}=15$ to $31.6\%$ at $\mathsf{op}=23$
(Figure~\ref{fig:pvc-vs-op}), and at $\mathsf{op}=23$ half of those traces
pass every syntactic and arithmetic check
(Table~\ref{tab:invalid-correct-causes}). Thus, under a reasoning-length
distribution shift, traces can retain the form of a plausible derivation
while increasingly losing their problem-specific semantics.

This failure mode is relevant to trace interpretation and CoT-based
oversight~\citep{korbak2025monitorability,guan2025monitoring,shah2026reasoning}.
Note that our experiments do not measure causal faithfulness or the accuracy of a safety
monitor, and we do not dispute that CoT monitoring can remain useful even when traces are imperfect. Our point is that an observable prerequisite for semantic interpretation---that the trace is valid for the problem it purports to solve---can degrade under distribution shift, even in an ideal setting: clean supervision, no adversary, and no incentive to conceal anything. Hence, correct outputs with highly plausible-looking traces alone are insufficient evidence that the semantics of the trace are trustworthy. Therefore, assuming that any reliability established on familiar tasks will transfer over to harder or unfamiliar ones, is unjustified in our view. 

Taken together, these experiments go beyond establishing answer--trace
decoupling. Within iGSM, a benchmark designed to study and
furnish evidence for claims of learned reasoning and planning
~\citep{ye2024physics21,ye2024physics22}, they test which properties of the visible trace are tied to successful computation and which can instead arise from supervision and problem construction. 
The results suggest that trace semantics matter for generalization, but useful intermediate computation need not coincide token-for-token with a correct, externally interpretable derivation.
The broader methodological lesson is that generated traces are
\emph{observations of model behavior}, not direct measurements of the reasoning process inside the model. Their validity, minimality, coherence, and
problem-specific meaning can be shaped by supervision and problem construction,
and their relationship to the final answer can change under distribution shift.
Inferring a reasoning mechanism from these properties therefore requires
counterfactual tests, not observation alone.

\section{Conclusion}
We have shown that in iGSM, where thinking traces are mechanically
verifiable, correct answers can come with invalid traces even when the model is trained exclusively on valid,
minimal traces. The fraction of correct answers accompanied by invalid traces
rises from below $1\%$ in distribution to $31.6\%$ on the hardest problems,
and over half of those invalid traces pass all syntactic and arithmetic checks.
Minimality is strongly shaped by supervision and query placement, so minimal
traces alone do not establish planning; meanwhile, training on swapped or
shuffled traces can preserve substantial answer accuracy despite producing no
valid traces. Our models are small and synthetic, so direct extrapolation to
frontier systems is unwarranted, but the simplicity and exact verifiability of
iGSM make these failures harder to dismiss, not easier. Whether thinking traces
of frontier models remain reliable under distribution shift must be tested
directly, not just inferred from correct answers and plausible-looking traces.

Appendix~\ref{sec:further-related} extends the related work;
Appendix~\ref{sec:experiment-details} gives the training and evaluation
details, the other clean runs, and a second seed of the swapped, shuffled
and non-minimal models; Appendix~\ref{sec:additional-tables} gives per-op
accuracies and the full-level counts behind Figure~\ref{fig:pvc-vs-op} and
Table~\ref{tab:invalid-correct-causes}; Appendix~\ref{sec:examples} gives
fifteen annotated example traces, one for each row of
Table~\ref{tab:invalid-correct-causes-full}.

\section*{Acknowledgements}
We acknowledge the EuroHPC Joint Undertaking for awarding this project
access to the EuroHPC supercomputer LEONARDO, hosted by CINECA (Italy) and
the LEONARDO consortium through an EuroHPC Development Access call. We
acknowledge the Danish e-Infrastructure Consortium (DeiC), Denmark, for
awarding this project access to the LUMI supercomputer, owned by the
EuroHPC Joint Undertaking, hosted by CSC (Finland) and the LUMI consortium
through DeiC, Denmark. Part of the computation was performed on Gefion,
operated by the Danish Centre for AI Innovation.
Kambhampati's research at Arizona State University is supported in part by grants from ONR grant N00014-25-1-2301, ARO grant W911NF-26-1-A278, and gifts from Qualcomm and Apple. 

\section*{AI use statement}
In this work, we used generative AI tools (Claude, Anthropic; ChatGPT and
Codex, OpenAI) to implement methods, including training, evaluation, and analysis.
We also used them for drafting and editing
sections of the paper, to search for related works, and for formatting references. 
We have not used generative AI tools to propose the research
questions, hypotheses or theoretical frameworks, to design the experiments or methodology,
to generate data sets (all problems come from the iGSM generator of
\citet{ye2024physics21}), or to interpret results; mathematical claims and
proofs, translation, data cleaning and qualitative analysis are not
applicable to this work. We have reviewed all AI-assisted work. We take
responsibility for the final content of this work, including text, claims
or artifacts produced with the aid of generative AI.

\section*{Reproducibility statement}
Code and trained models are available at \url{https://github.com/ratishsp/igsm-trace-validity} and \url{https://huggingface.co/ratishsp/igsm-trace-validity}.

\bibliography{iclr2027_conference}

\begin{thebibliography}{41}
\providecommand{\natexlab}[1]{#1}
\providecommand{\url}[1]{\texttt{#1}}
\expandafter\ifx\csname urlstyle\endcsname\relax
  \providecommand{\doi}[1]{doi: #1}\else
  \providecommand{\doi}{doi: \begingroup \urlstyle{rm}\Url}\fi

\bibitem[Baker et~al.(2025)Baker, Huizinga, Gao, Dou, Guan, Madry, Zaremba,
  Pachocki, and Farhi]{baker2025monitoring}
Bowen Baker, Joost Huizinga, Leo Gao, Zehao Dou, Melody~Y. Guan, Aleksander
  Madry, Wojciech Zaremba, Jakub Pachocki, and David Farhi.
\newblock Monitoring reasoning models for misbehavior and the risks of
  promoting obfuscation.
\newblock \emph{arXiv preprint arXiv:2503.11926}, 2025.
\newblock URL \url{https://arxiv.org/abs/2503.11926}.

\bibitem[Bhambri et~al.(2026)Bhambri, Biswas, and
  Kambhampati]{bhambri2026interpretable}
Siddhant Bhambri, Upasana Biswas, and Subbarao Kambhampati.
\newblock Interpretable traces, unexpected outcomes: Investigating the
  disconnect in trace-based knowledge distillation.
\newblock In \emph{Proceedings of the 64th Annual Meeting of the Association
  for Computational Linguistics (Volume 1: Long Papers)}, pp.\  36390--36408,
  2026.
\newblock \doi{10.18653/v1/2026.acl-long.1686}.
\newblock URL \url{https://aclanthology.org/2026.acl-long.1686/}.

\bibitem[Brown-Cohen et~al.(2026)Brown-Cohen, Lindner, and
  Shah]{browncohen2026opaque}
Jonah Brown-Cohen, David Lindner, and Rohin Shah.
\newblock Quantifying the necessity of chain of thought through opaque serial
  depth.
\newblock \emph{arXiv preprint arXiv:2603.09786}, 2026.
\newblock URL \url{https://arxiv.org/abs/2603.09786}.

\bibitem[Chen et~al.(2024)Chen, Su, Zuo, Yang, Yuan, Chan, Yu, Lu, Hung, Qian,
  Qin, Cong, Xie, Liu, Sun, and Zhou]{chen2024agentverse}
Weize Chen, Yusheng Su, Jingwei Zuo, Cheng Yang, Chenfei Yuan, Chi-Min Chan,
  Heyang Yu, Yaxi Lu, Yi-Hsin Hung, Chen Qian, Yujia Qin, Xin Cong, Ruobing
  Xie, Zhiyuan Liu, Maosong Sun, and Jie Zhou.
\newblock {AgentVerse}: Facilitating multi-agent collaboration and exploring
  emergent behaviors.
\newblock In \emph{The Twelfth International Conference on Learning
  Representations}, 2024.
\newblock URL
  \url{https://proceedings.iclr.cc/paper_files/paper/2024/hash/578e65cdee35d00c708d4c64bce32971-Abstract-Conference.html}.

\bibitem[Chen et~al.(2025)Chen, Benton, Radhakrishnan, Uesato, Denison,
  Schulman, Somani, Hase, Wagner, Roger, Mikulik, Bowman, Leike, Kaplan, and
  Perez]{chen2025reasoning}
Yanda Chen, Joe Benton, Ansh Radhakrishnan, Jonathan Uesato, Carson Denison,
  John Schulman, Arushi Somani, Peter Hase, Misha Wagner, Fabien Roger, Vlad
  Mikulik, Samuel~R. Bowman, Jan Leike, Jared Kaplan, and Ethan Perez.
\newblock Reasoning models don't always say what they think.
\newblock \emph{arXiv preprint arXiv:2505.05410}, 2025.
\newblock URL \url{https://arxiv.org/abs/2505.05410}.

\bibitem[Cobbe et~al.(2021)Cobbe, Kosaraju, Bavarian, Chen, Jun, Kaiser,
  Plappert, Tworek, Hilton, Nakano, Hesse, and Schulman]{cobbe2021training}
Karl Cobbe, Vineet Kosaraju, Mohammad Bavarian, Mark Chen, Heewoo Jun, Lukasz
  Kaiser, Matthias Plappert, Jerry Tworek, Jacob Hilton, Reiichiro Nakano,
  Christopher Hesse, and John Schulman.
\newblock Training verifiers to solve math word problems.
\newblock \emph{arXiv preprint arXiv:2110.14168}, 2021.
\newblock URL \url{https://arxiv.org/abs/2110.14168}.

\bibitem[Emmons et~al.(2025)Emmons, Jenner, Elson, Saurous, Rajamanoharan,
  Chen, Shafkat, and Shah]{emmons2025necessary}
Scott Emmons, Erik Jenner, David~K. Elson, Rif~A. Saurous, Senthooran
  Rajamanoharan, Heng Chen, Irhum Shafkat, and Rohin Shah.
\newblock When chain of thought is necessary, language models struggle to evade
  monitors.
\newblock \emph{arXiv preprint arXiv:2507.05246}, 2025.
\newblock URL \url{https://arxiv.org/abs/2507.05246}.

\bibitem[Goyal et~al.(2024)Goyal, Ji, Rawat, Menon, Kumar, and
  Nagarajan]{goyal2024think}
Sachin Goyal, Ziwei Ji, Ankit~Singh Rawat, Aditya~Krishna Menon, Sanjiv Kumar,
  and Vaishnavh Nagarajan.
\newblock Think before you speak: Training language models with pause tokens.
\newblock In \emph{International Conference on Learning Representations}, 2024.
\newblock URL
  \url{https://proceedings.iclr.cc/paper_files/paper/2024/hash/76917808731dae9e6d62c2a7a6afb542-Abstract-Conference.html}.

\bibitem[Greenblatt et~al.(2026)Greenblatt, Cotra, and
  Wijk]{metr2026huggingface}
Ryan Greenblatt, Ajeya Cotra, and Hjalmar Wijk.
\newblock Brief independent investigation of agents' behavior, reasoning and
  collaboration in the {OpenAI} / {Hugging Face} hacking incident.
\newblock METR and Redwood Research, August 2026.
\newblock URL
  \url{https://metr.org/blog/2026-08-26-openai-hugging-face-incident-investigation/}.
\newblock Published August 26, 2026.

\bibitem[Guan et~al.(2026)Guan, Wang, Carroll, Dou, Wei, Williams, Arnav,
  Huizinga, Kivlichan, Glaese, Pachocki, and Baker]{guan2025monitoring}
Melody~Y. Guan, Miles Wang, Micah Carroll, Zehao Dou, Annie~Y. Wei, Marcus
  Williams, Benjamin Arnav, Joost Huizinga, Ian Kivlichan, Amelia Glaese, Jakub
  Pachocki, and Bowen Baker.
\newblock Monitoring monitorability.
\newblock In \emph{Proceedings of the 43rd International Conference on Machine
  Learning}, 2026.
\newblock URL \url{https://openreview.net/forum?id=b82fgbMVpz}.

\bibitem[Guo et~al.(2025)Guo, Yang, Zhang, Song, Wang, Zhu, Xu, Zhang, Ma, Bi,
  et~al.]{deepseekai2025r1}
Daya Guo, Dejian Yang, Haowei Zhang, Junxiao Song, Peiyi Wang, Qihao Zhu,
  Runxin Xu, Ruoyu Zhang, Shirong Ma, Xiao Bi, et~al.
\newblock {DeepSeek-R1} incentivizes reasoning in {LLM}s through reinforcement
  learning.
\newblock \emph{Nature}, 645:\penalty0 633--638, 2025.
\newblock \doi{10.1038/s41586-025-09422-z}.
\newblock URL \url{https://doi.org/10.1038/s41586-025-09422-z}.

\bibitem[Haskins et~al.(2026)Haskins, Chughtai, and
  Engels]{haskins2026training}
Reilly Haskins, Bilal Chughtai, and Joshua Engels.
\newblock Training on documents about monitoring leads to {CoT} obfuscation.
\newblock \emph{arXiv preprint arXiv:2605.15257}, 2026.
\newblock URL \url{https://arxiv.org/abs/2605.15257}.

\bibitem[Kambhampati et~al.(2026)Kambhampati, Valmeekam, Bhambri, Palod,
  Saldyt, Stechly, Samineni, Kalwar, and Biswas]{kambhampati2025stop}
Subbarao Kambhampati, Karthik Valmeekam, Siddhant Bhambri, Vardhan Palod, Lucas
  Saldyt, Kaya Stechly, Soumya~Rani Samineni, Durgesh Kalwar, and Upasana
  Biswas.
\newblock Position: Stop anthropomorphizing intermediate tokens as
  reasoning/thinking traces!
\newblock In \emph{Proceedings of the 43rd International Conference on Machine
  Learning}, 2026.
\newblock URL \url{https://openreview.net/forum?id=nP7rL36vYj}.

\bibitem[{Kimi Team} et~al.(2025)]{kimi2025k15}
{Kimi Team} et~al.
\newblock {Kimi k1.5}: Scaling reinforcement learning with {LLM}s.
\newblock \emph{arXiv preprint arXiv:2501.12599}, 2025.
\newblock URL \url{https://arxiv.org/abs/2501.12599}.

\bibitem[Kojima et~al.(2022)Kojima, Gu, Reid, Matsuo, and
  Iwasawa]{kojima2022large}
Takeshi Kojima, Shixiang~Shane Gu, Machel Reid, Yutaka Matsuo, and Yusuke
  Iwasawa.
\newblock Large language models are zero-shot reasoners.
\newblock In \emph{Advances in Neural Information Processing Systems},
  volume~35, pp.\  22199--22213, 2022.
\newblock \doi{10.52202/068431-1613}.
\newblock URL
  \url{https://proceedings.neurips.cc/paper/2022/hash/8bb0d291acd4acf06ef112099c16f326-Abstract-Conference.html}.

\bibitem[Korbak et~al.(2025)Korbak, Balesni, Barnes, Bengio, Benton, Bloom,
  Chen, Cooney, Dafoe, Dragan, Emmons, Evans, Farhi, Greenblatt, Hendrycks,
  Hobbhahn, Hubinger, Irving, Jenner, Kokotajlo, Krakovna, Legg, Lindner, Luan,
  Madry, Michael, Nanda, Orr, Pachocki, Perez, Phuong, Roger, Saxe, Shlegeris,
  Soto, Steinberger, Wang, Zaremba, Baker, Shah, and
  Mikulik]{korbak2025monitorability}
Tomek Korbak, Mikita Balesni, Elizabeth Barnes, Yoshua Bengio, Joe Benton,
  Joseph Bloom, Mark Chen, Alan Cooney, Allan Dafoe, Anca Dragan, Scott Emmons,
  Owain Evans, David Farhi, Ryan Greenblatt, Dan Hendrycks, Marius Hobbhahn,
  Evan Hubinger, Geoffrey Irving, Erik Jenner, Daniel Kokotajlo, Victoria
  Krakovna, Shane Legg, David Lindner, David Luan, Aleksander Madry, Julian
  Michael, Neel Nanda, Dave Orr, Jakub Pachocki, Ethan Perez, Mary Phuong,
  Fabien Roger, Joshua Saxe, Buck Shlegeris, Mart{\'\i}n Soto, Eric
  Steinberger, Jasmine Wang, Wojciech Zaremba, Bowen Baker, Rohin Shah, and
  Vlad Mikulik.
\newblock Chain of thought monitorability: A new and fragile opportunity for
  {AI} safety.
\newblock \emph{arXiv preprint arXiv:2507.11473}, 2025.
\newblock URL \url{https://arxiv.org/abs/2507.11473}.

\bibitem[Lanham et~al.(2023)Lanham, Chen, Radhakrishnan, Steiner, Denison,
  Hernandez, Li, Durmus, Hubinger, Kernion, Luko{\v{s}}i{\={u}}t{\.{e}},
  Nguyen, Cheng, Joseph, Schiefer, Rausch, Larson, McCandlish, Kundu, Kadavath,
  Yang, Henighan, Maxwell, Telleen-Lawton, Hume, Hatfield-Dodds, Kaplan,
  Brauner, Bowman, and Perez]{lanham2023measuring}
Tamera Lanham, Anna Chen, Ansh Radhakrishnan, Benoit Steiner, Carson Denison,
  Danny Hernandez, Dustin Li, Esin Durmus, Evan Hubinger, Jackson Kernion,
  Kamil{\.{e}} Luko{\v{s}}i{\={u}}t{\.{e}}, Karina Nguyen, Newton Cheng,
  Nicholas Joseph, Nicholas Schiefer, Oliver Rausch, Robin Larson, Sam
  McCandlish, Sandipan Kundu, Saurav Kadavath, Shannon Yang, Thomas Henighan,
  Timothy Maxwell, Timothy Telleen-Lawton, Tristan Hume, Zac Hatfield-Dodds,
  Jared Kaplan, Jan Brauner, Samuel~R. Bowman, and Ethan Perez.
\newblock Measuring faithfulness in chain-of-thought reasoning.
\newblock \emph{arXiv preprint arXiv:2307.13702}, 2023.
\newblock URL \url{https://arxiv.org/abs/2307.13702}.

\bibitem[Li et~al.(2023)Li, Hammoud, Itani, Khizbullin, and
  Ghanem]{li2023camel}
Guohao Li, Hasan Abed Al~Kader Hammoud, Hani Itani, Dmitrii Khizbullin, and
  Bernard Ghanem.
\newblock {CAMEL}: Communicative agents for ``mind'' exploration of large
  language model society.
\newblock In \emph{Advances in Neural Information Processing Systems},
  volume~36, 2023.
\newblock \doi{10.52202/075280-2264}.
\newblock URL
  \url{https://proceedings.neurips.cc/paper_files/paper/2023/hash/a3621ee907def47c1b952ade25c67698-Abstract-Conference.html}.

\bibitem[Lightman et~al.(2024)Lightman, Kosaraju, Burda, Edwards, Baker, Lee,
  Leike, Schulman, Sutskever, and Cobbe]{lightman2023verify}
Hunter Lightman, Vineet Kosaraju, Yuri Burda, Harrison Edwards, Bowen Baker,
  Teddy Lee, Jan Leike, John Schulman, Ilya Sutskever, and Karl Cobbe.
\newblock Let's verify step by step.
\newblock In \emph{The Twelfth International Conference on Learning
  Representations}, 2024.
\newblock URL \url{https://openreview.net/forum?id=v8L0pN6EOi}.

\bibitem[Lyons(2026)]{lyons2026agentcollective}
Jessica Lyons.
\newblock Openai reveals its rogue agent swarm went a little bit {Borg} ahead
  of hugging face hack.
\newblock \emph{The Register}, August 2026.
\newblock URL
  \url{https://www.theregister.com/security/2026/08/06/openai-reveals-its-rogue-agent-swarm-went-a-little-bit-borg-ahead-of-hugging-face-hack/5283741}.

\bibitem[McCarthy et~al.(2025)McCarthy, Skaf, Ibanez-Lissen, Georgiev, Watts,
  Whittingham, Gonzalez-Manzano, Tice, Young, Radmard, and
  Lindner]{mccarthy2025steganographic}
Robert McCarthy, Joey Skaf, Luis Ibanez-Lissen, Vasil Georgiev, Connor Watts,
  Hannes Whittingham, Lorena Gonzalez-Manzano, Cameron Tice, Edward~James
  Young, Puria Radmard, and David Lindner.
\newblock Large language models can learn and generalize steganographic
  chain-of-thought under process supervision.
\newblock In \emph{Advances in Neural Information Processing Systems},
  volume~38, 2025.
\newblock \doi{10.52202/085713-0930}.
\newblock URL
  \url{https://papers.neurips.cc/paper_files/paper/2025/hash/28131b22fafebba500eb7bb02e3d5b59-Abstract-Conference.html}.

\bibitem[Mirzadeh et~al.(2025)Mirzadeh, Alizadeh-Vahid, Shahrokhi, Tuzel,
  Bengio, and Farajtabar]{mirzadeh2025gsm}
Iman Mirzadeh, Keivan Alizadeh-Vahid, Hooman Shahrokhi, Oncel Tuzel, Samy
  Bengio, and Mehrdad Farajtabar.
\newblock {GSM}-symbolic: Understanding the limitations of mathematical
  reasoning in large language models.
\newblock In \emph{International Conference on Learning Representations}, 2025.
\newblock URL
  \url{https://proceedings.iclr.cc/paper_files/paper/2025/hash/ec2e7a896f8250986b3907f57621ce94-Abstract-Conference.html}.

\bibitem[Pfau et~al.(2024)Pfau, Merrill, and Bowman]{pfau2024dot}
Jacob Pfau, William Merrill, and Samuel~R. Bowman.
\newblock Let's think dot by dot: Hidden computation in transformer language
  models.
\newblock In \emph{First Conference on Language Modeling}, 2024.
\newblock URL \url{https://openreview.net/forum?id=NikbrdtYvG}.

\bibitem[Radford et~al.(2019)Radford, Wu, Child, Luan, Amodei, and
  Sutskever]{radford2019language}
Alec Radford, Jeffrey Wu, Rewon Child, David Luan, Dario Amodei, and Ilya
  Sutskever.
\newblock Language models are unsupervised multitask learners.
\newblock Technical report, OpenAI, 2019.
\newblock URL
  \url{https://cdn.openai.com/better-language-models/language_models_are_unsupervised_multitask_learners.pdf}.

\bibitem[Shah \& Dragan(2026)Shah and Dragan]{shah2026reasoning}
Rohin Shah and Anca Dragan.
\newblock The case for reasoning transparency.
\newblock DeepMind Institute essay, September 2026.
\newblock URL
  \url{https://institute.deepmind.com/essays/the-case-for-reasoning-transparency/}.

\bibitem[Shao et~al.(2024)Shao, Wang, Zhu, Xu, Song, Bi, Zhang, Zhang, Li, Wu,
  and Guo]{shao2024deepseekmath}
Zhihong Shao, Peiyi Wang, Qihao Zhu, Runxin Xu, Junxiao Song, Xiao Bi, Haowei
  Zhang, Mingchuan Zhang, Y.~K. Li, Y.~Wu, and Daya Guo.
\newblock Deepseekmath: Pushing the limits of mathematical reasoning in open
  language models.
\newblock \emph{arXiv preprint arXiv:2402.03300}, 2024.
\newblock URL \url{https://arxiv.org/abs/2402.03300}.

\bibitem[Su et~al.(2024)Su, Ahmed, Lu, Pan, Bo, and Liu]{su2024roformer}
Jianlin Su, Murtadha Ahmed, Yu~Lu, Shengfeng Pan, Wen Bo, and Yunfeng Liu.
\newblock {RoFormer}: Enhanced transformer with rotary position embedding.
\newblock \emph{Neurocomputing}, 568:\penalty0 127063, 2024.
\newblock \doi{10.1016/j.neucom.2023.127063}.

\bibitem[Turpin et~al.(2023)Turpin, Michael, Perez, and
  Bowman]{turpin2023language}
Miles Turpin, Julian Michael, Ethan Perez, and Samuel~R. Bowman.
\newblock Language models don't always say what they think: Unfaithful
  explanations in chain-of-thought prompting.
\newblock In \emph{Advances in Neural Information Processing Systems},
  volume~36, pp.\  74952--74965, 2023.
\newblock \doi{10.52202/075280-3275}.
\newblock URL
  \url{https://proceedings.neurips.cc/paper_files/paper/2023/hash/ed3fea9033a80fea1376299fa7863f4a-Abstract-Conference.html}.

\bibitem[Uesato et~al.(2022)Uesato, Kushman, Kumar, Song, Siegel, Wang,
  Creswell, Irving, and Higgins]{uesato2022solving}
Jonathan Uesato, Nate Kushman, Ramana Kumar, Francis Song, Noah Siegel, Lisa
  Wang, Antonia Creswell, Geoffrey Irving, and Irina Higgins.
\newblock Solving math word problems with process- and outcome-based feedback.
\newblock \emph{arXiv preprint arXiv:2211.14275}, 2022.
\newblock URL \url{https://arxiv.org/abs/2211.14275}.

\bibitem[Valmeekam et~al.(2026)Valmeekam, Palod, Stechly, Gundawar, and
  Kambhampati]{valmeekam2026beyond}
Karthik Valmeekam, Vardhan Palod, Kaya Stechly, Atharva Gundawar, and Subbarao
  Kambhampati.
\newblock Beyond semantics: The unreasonable effectiveness of reasonless
  intermediate tokens.
\newblock \emph{Transactions on Machine Learning Research}, 2026.
\newblock URL \url{https://openreview.net/forum?id=gDE7YcRC3F}.

\bibitem[Wang et~al.(2023)Wang, Min, Deng, Shen, Wu, Zettlemoyer, and
  Sun]{wang2023towards}
Boshi Wang, Sewon Min, Xiang Deng, Jiaming Shen, You Wu, Luke Zettlemoyer, and
  Huan Sun.
\newblock Towards understanding chain-of-thought prompting: An empirical study
  of what matters.
\newblock In \emph{Proceedings of the 61st Annual Meeting of the Association
  for Computational Linguistics (Volume 1: Long Papers)}, pp.\  2717--2739.
  Association for Computational Linguistics, 2023.
\newblock \doi{10.18653/v1/2023.acl-long.153}.
\newblock URL \url{https://aclanthology.org/2023.acl-long.153/}.

\bibitem[Wang et~al.(2024)Wang, Li, Shao, Xu, Dai, Li, Chen, Wu, and
  Sui]{wang2024mathshepherd}
Peiyi Wang, Lei Li, Zhihong Shao, Runxin Xu, Damai Dai, Yifei Li, Deli Chen,
  Yu~Wu, and Zhifang Sui.
\newblock Math-shepherd: Verify and reinforce {LLM}s step-by-step without human
  annotations.
\newblock In \emph{Proceedings of the 62nd Annual Meeting of the Association
  for Computational Linguistics (Volume 1: Long Papers)}, pp.\  9426--9439,
  Bangkok, Thailand, 2024. Association for Computational Linguistics.
\newblock \doi{10.18653/v1/2024.acl-long.510}.
\newblock URL \url{https://aclanthology.org/2024.acl-long.510/}.

\bibitem[Wei et~al.(2022)Wei, Wang, Schuurmans, Bosma, Ichter, Xia, Chi, Le,
  and Zhou]{wei2022chain}
Jason Wei, Xuezhi Wang, Dale Schuurmans, Maarten Bosma, Brian Ichter, Fei Xia,
  Ed~H. Chi, Quoc~V. Le, and Denny Zhou.
\newblock Chain-of-thought prompting elicits reasoning in large language
  models.
\newblock In \emph{Advances in Neural Information Processing Systems},
  volume~35, pp.\  24824--24837, 2022.
\newblock \doi{10.52202/068431-1800}.
\newblock URL
  \url{https://proceedings.neurips.cc/paper_files/paper/2022/hash/9d5609613524ecf4f15af0f7b31abca4-Abstract-Conference.html}.

\bibitem[Wong(2026)]{wong2026panic}
Matteo Wong.
\newblock It may be time to panic about {AI}.
\newblock \emph{The Atlantic}, August 2026.
\newblock URL
  \url{https://www.theatlantic.com/technology/2026/08/openai-hacks-panic/688264/}.

\bibitem[Wu et~al.(2024)Wu, Bansal, Zhang, Wu, Li, Zhu, Jiang, Zhang, Zhang,
  Liu, Awadallah, White, Burger, and Wang]{wu2024autogen}
Qingyun Wu, Gagan Bansal, Jieyu Zhang, Yiran Wu, Beibin Li, Erkang Zhu,
  Li~Jiang, Xiaoyun Zhang, Shaokun Zhang, Jiale Liu, Ahmed~Hassan Awadallah,
  Ryen~W. White, Doug Burger, and Chi Wang.
\newblock {AutoGen}: Enabling next-gen {LLM} applications via multi-agent
  conversations.
\newblock In \emph{First Conference on Language Modeling}, 2024.
\newblock URL \url{https://openreview.net/forum?id=BAakY1hNKS}.

\bibitem[Yao et~al.(2026)Yao, Chen, and Zhang]{yao2026competition}
Jiayi Yao, Cong Chen, and Baosen Zhang.
\newblock Competition and cooperation of {LLM} agents in games, 2026.
\newblock URL \url{https://arxiv.org/abs/2604.00487}.
\newblock arXiv preprint.

\bibitem[Ye et~al.(2025{\natexlab{a}})Ye, Xu, Li, and
  {Allen-Zhu}]{ye2024physics21}
Tian Ye, Zicheng Xu, Yuanzhi Li, and Zeyuan {Allen-Zhu}.
\newblock Physics of language models: Part 2.1, grade-school math and the
  hidden reasoning process.
\newblock In \emph{The Thirteenth International Conference on Learning
  Representations}, 2025{\natexlab{a}}.
\newblock URL
  \url{https://proceedings.iclr.cc/paper_files/paper/2025/hash/f3064f7a0ca2328ecb41a3aef6177d68-Abstract-Conference.html}.

\bibitem[Ye et~al.(2025{\natexlab{b}})Ye, Xu, Li, and
  {Allen-Zhu}]{ye2024physics22}
Tian Ye, Zicheng Xu, Yuanzhi Li, and Zeyuan {Allen-Zhu}.
\newblock Physics of language models: Part 2.2, how to learn from mistakes on
  grade-school math problems.
\newblock In \emph{The Thirteenth International Conference on Learning
  Representations}, 2025{\natexlab{b}}.
\newblock URL
  \url{https://proceedings.iclr.cc/paper_files/paper/2025/hash/c239bac713017b0b2257b7622bf8aab3-Abstract-Conference.html}.

\bibitem[Zelikman et~al.(2022)Zelikman, Wu, Mu, and Goodman]{zelikman2022star}
Eric Zelikman, Yuhuai Wu, Jesse Mu, and Noah~D. Goodman.
\newblock {STaR}: Bootstrapping reasoning with reasoning.
\newblock In \emph{Advances in Neural Information Processing Systems},
  volume~35, pp.\  15476--15488, 2022.
\newblock \doi{10.52202/068431-1126}.
\newblock URL
  \url{https://proceedings.neurips.cc/paper_files/paper/2022/hash/639a9a172c044fbb64175b5fad42e9a5-Abstract-Conference.html}.

\bibitem[Zhou et~al.(2025)Zhou, Liu, Chen, Tian, and Chen]{zhou2025gsminf}
Yang Zhou, Hongyi Liu, Zhuoming Chen, Yuandong Tian, and Beidi Chen.
\newblock {GSM}-$\infty$: How do your {LLM}s behave over infinitely increasing
  reasoning complexity and context length?
\newblock In \emph{Proceedings of the 42nd International Conference on Machine
  Learning}, volume 267 of \emph{Proceedings of Machine Learning Research},
  pp.\  78933--78983, 2025.
\newblock URL \url{https://proceedings.mlr.press/v267/zhou25m.html}.

\bibitem[Zolkowski et~al.(2025)Zolkowski, Xing, Lindner, Tramèr, and
  Jenner]{zolkowski2025obfuscate}
Artur Zolkowski, Wen Xing, David Lindner, Florian Tramèr, and Erik Jenner.
\newblock Can reasoning models obfuscate reasoning? stress-testing
  chain-of-thought monitorability.
\newblock \emph{arXiv preprint arXiv:2510.19851}, 2025.
\newblock URL \url{https://arxiv.org/abs/2510.19851}.

\end{thebibliography}
\bibliographystyle{iclr2027_conference}

\clearpage
\appendix
\section{Further Related Work}
\label{sec:further-related}
\paragraph{Traces, chain of thought and filler tokens.}
Extra tokens between a question and its answer can help even when they carry no meaning. \citet{goyal2024think} add a learnable pause token and find gains when it is used in both pretraining and finetuning, though the number of pause tokens has to be tuned per task. \citet{pfau2024dot} replace chain of thought with repeated filler tokens and show that on a synthetic parallelizable task like 3SUM there is a minimal loss of performance over parallelizable CoT. However, filler training does not seem to recover the behaviour of standard, serial CoT. 
With prompting, demonstrations whose reasoning steps are wrong keep most of the benefit of correct ones, as long as the steps stay relevant and in a sensible order \citep{wang2023towards}.
Truncating or corrupting a model's own trace often leaves its answer unchanged, though how often depends on the task \citep{lanham2023measuring}.
\citet{kambhampati2025stop} argue from results like these that intermediate tokens should not be read as thoughts. Our swapped traces fall between filler and real chain of thought. They are well formed and look like solutions, but they belong to a different problem.

\paragraph{Verifying intermediate steps.}
Most of the work above judges traces by their effect on the answer, since natural-language traces cannot be reliably checked by a program, requiring human judgment or a learned verifier. \citet{uesato2022solving} had humans label errors in GSM8K solutions and found that outcome supervision gives about the same final-answer error as process supervision, but more correct answers reached through wrong steps. 
\citet{lightman2023verify} make a similar point on MATH, where grading by the final answer alone counts a solution as correct when the answer is right and the steps are wrong.
\citet{bhambri2026interpretable} obtain the same separation without human labels. Working in question answering with a rule-based problem decomposition, they build fine-tuning sets in which each problem is paired with either a verifiably correct or a verifiably incorrect CoT trace while the final solution is always correct. Correct traces lead to correct solutions on only $28\%$ of test problems, and incorrect traces do not necessarily degrade solution accuracy, so trace correctness and answer correctness are not reliably coupled. They also study end-user interpretability, where the traces that produce the best models are not the ones readers find most interpretable.
\citet{valmeekam2026beyond} demonstrated a striking separation between the semantics of reasoning tokens and the correctness of the final output for maze solving.
In that setting, the intended reasoning trace corresponds to the execution of a formal search algorithm and can therefore be checked mechanically. They show that models could produce correct plans despite invalid traces. Moreover, training on completely unrelated traces still retains much of the benefit of training on valid traces. 
Our study follows their setup, but in grade-school math, where the traces are derivations rather than search logs, and where prior work has argued that models trained on this data reason through their traces \citep{ye2024physics21}.

\paragraph{Generalizing to longer and harder problems.}
\citet{ye2024physics21} train GPT-2 models on iGSM problems with up to 15 or 21 operations and find that they stay accurate on problems that need more operations than any seen in training. They take this as evidence that the model learned to reason rather than memorize templates. Their accuracy counts a solution only if every step is valid, so it does not tell us how answer correctness and trace validity relate as the problems get harder. On general pretrained language models, there is a notable loss in performance compared to the previous setup. Finetuning on GSM8K variants with one extra clause does not transfer to variants with two \citep{mirzadeh2025gsm}, and accuracy on GSM-Infinite falls off along a sigmoid as the number of operations grows \citep{zhou2025gsminf}. In the problem of solving mazes, the swapped-trace models of \citet{valmeekam2026beyond} generalize well, and sometimes best, on out-of-distribution maze types. 
In our setting, where difficulty grows with the number of operations, answer accuracy of the swapped-trace model drops sharply relative to models trained on correct traces as the number of operations grows. One explanation is that the out-of-distribution mazes change the structure within a fixed grid, whereas harder iGSM problems require more serial computation because they involve more operations.
This is consistent with the view that trace content becomes more important as the required computation exceeds what can be captured in a single forward pass \citep{emmons2025necessary}.

\paragraph{Grade-school math.}
iGSM \citep{ye2024physics21} generates problems from a hierarchy of categories and a random dependency graph, with arithmetic mod 23, so every solution can be checked step by step by a program. GSM-Infinite builds on a similar graph construction, reworks the noisy data generator, drops the modulus, adds division and a reverse mode, and scales the number of operations and the context length for evaluating pretrained models \citep{zhou2025gsminf}. GSM-Symbolic turns GSM8K test problems into templates and finds large variance across instances, and big drops when an irrelevant clause is added \citep{mirzadeh2025gsm}. Both benchmarks score only the final answer. GSM-Symbolic traces are free-form and cannot be checked by a program reliably. GSM-Infinite has the graphs needed to check them, but does not use them for that. Given our results, accuracy curves on these benchmarks may not say much about whether the reasoning behind them is correct.

\paragraph{Chain of thought in AI safety.}
Reading a model's chain of thought has become a key tool for oversight.
\citet{korbak2025monitorability} and \citet{shah2026reasoning} argue that this window is valuable but fragile, and that its value rests on the trace having been optimized to be useful for the answer, not to look a particular way. 
\citet{guan2025monitoring} build evaluations to track monitorability across models and training, and find that most current frontier models are fairly, but not perfectly, monitorable. Other work shows how monitorability can break.
Penalizing bad reasoning in the trace during RL leads to reward hacking with benign-looking traces \citep{baker2025monitoring}, penalizing specific strings leads to encoded reasoning \citep{mccarthy2025steganographic}, and models finetuned only on documents describing a monitor learn to leave their misbehavior out of the trace \citep{haskins2026training}.
Some prompted models can also hide a side task from a monitor under enough pressure \citep{zolkowski2025obfuscate}. 
\citet{emmons2025necessary} find that models
struggle to evade monitors on tasks that require chain of thought, and \citet{browncohen2026opaque} bound how much serial computation a transformer can do without writing it down. 
Our setting has no adversary, no monitor in the loop and no pressure on the trace to be optimal. 
Even so, correct answers often come with invalid traces, more so on harder
problems. The answer accuracy of our swapped-trace models collapses only once problems get large, which is consistent with the necessity view. But a trace being necessary does not make
it correct, so a monitor can read a necessary trace and still see the wrong
derivation.

\section{Experiments and Results}
\label{sec:experiment-details}
\paragraph{Setup.}
We follow the iGSM-med recipe of \citet{ye2024physics21}. Every model is a
12-layer GPT-2-style transformer~\citep{radford2019language} with hidden
size 768 and rotary position embeddings~\citep{su2024roformer}, 124M
parameters, trained from scratch for 100k steps at batch size 512 on
problems with $\mathsf{op}\le 15$ and $\mathsf{ip}\le 20$ generated on the
fly. The models differ only in the trace that follows each problem in the
training data. We evaluate with greedy decoding on a fixed set of 4,096 problems per
level. The levels are in distribution at $\mathsf{op}\le 15$ and $\mathsf{op}=15$, and
out of distribution at $\mathsf{op}\in\{20,\dots,23\}$. Answers are integers modulo 23, so a random guess is
correct with probability $1/23 \approx 4.3\%$. However, because a product is $0$ whenever a factor is, a $0$ anywhere on the
chain propagates to every quantity computed from it via multiplication. This results in 14--16\% of problems having correct answer $0$.  A model that always answers $0$ therefore scores about 15\%, and accuracies near that value are not statistically significant. A trace is \emph{valid} when it passes every check of the
evaluator described in Section~\ref{sec:background}. 

We trained several clean models with this recipe. Their accuracies are in
Table~\ref{tab:clean-runs}, together with the values reported by
\citet{ye2024physics21}. Our reproductions fall short of their reported
values out of distribution.
Every table in the paper uses one of them, run A. Answer accuracy varies between
runs; two runs of the recipe differ by ten points at $\mathsf{op}=23$, so
we read differences of that size between models as noise. Doubling the training to 200k steps (run C) leaves the picture unchanged. At $\mathsf{op}=23$, answer accuracy moves from $58.0\%$ to $58.5\%$, and the share of instances with both a correct answer and a valid trace from $39.7\%$ to $41.3\%$. The share of correct answers with a valid trace at $\mathsf{op}=23$ is stable across runs, $68\%$ for run A, $68\%$ for the second run of the recipe (run B), and $71\%$ for run C, even though answer accuracy is not. A second training seed of each model trained on swapped, shuffled or non-minimal traces reproduces its accuracies within five points at every level, with the same ordering of the models (Table~\ref{tab:seed-replicates}).

\section{Additional tables}
\label{sec:additional-tables}
Tables~\ref{tab:arms-swapped-per-op} and~\ref{tab:arms-shuffled-per-op} give
the per-op accuracies behind Tables~\ref{tab:arms-swapped}
and~\ref{tab:arms-shuffled} respectively. Table~\ref{tab:quadrants-clean-full} gives the counts behind Figure~\ref{fig:pvc-vs-op} at all five op levels, while Table~\ref{tab:invalid-correct-causes-full} extends the failure breakdown in Table~\ref{tab:invalid-correct-causes}, using the same denominators. At $\mathsf{op}=15,20,23$, the shares of invalid traces accompanying correct answers that pass all shallow checks but fail semantic checks are $95.0\%$, $78.1\%$, and $51.2\%$ in both tables. The additional $\mathsf{op}=21,22$ columns report $63.2\%$ and $55.1\%$, respectively. Table~\ref{tab:clean-runs} compares the three
clean training runs, and Table~\ref{tab:seed-replicates} gives a second training seed for the models trained on swapped, shuffled and non-minimal traces.
\begin{table}[h]
\centering
\begin{tabular}{r r rrr}
\toprule
op & problems & clean & swapped & swapped, op-matched \\
\midrule
1 & 553 & 100.0 & 100.0 & 97.8 \\
2 & 493 & 100.0 & 100.0 & 98.6 \\
3 & 483 & 100.0 & 99.8 & 99.6 \\
4 & 419 & 100.0 & 100.0 & 100.0 \\
5 & 396 & 100.0 & 96.5 & 97.2 \\
6 & 325 & 100.0 & 89.2 & 91.4 \\
7 & 266 & 100.0 & 82.3 & 84.2 \\
8 & 259 & 100.0 & 66.4 & 67.6 \\
9 & 240 & 100.0 & 56.7 & 50.8 \\
10 & 194 & 100.0 & 36.6 & 44.3 \\
11 & 182 & 100.0 & 29.7 & 29.1 \\
12 & 127 & 100.0 & 33.9 & 27.6 \\
13 & 91 & 100.0 & 18.7 & 15.4 \\
14 & 55 & 98.2 & 16.4 & 20.0 \\
15 & 4096 & 98.8 & 19.9 & 19.2 \\
\bottomrule
\end{tabular}
\caption{Accuracies for iGSM with clean, swapped traces, and swapped traces with operations matched, for each in-distribution operation count.}
\label{tab:arms-swapped-per-op}
\end{table}

\begin{table}[h]
\centering
\begin{tabular}{r r r rrrrr rr}
\toprule
op & problems & clean & 10\% & 30\% & 50\% & 75\% & 100\% & swapped & no trace \\
\midrule
1 & 553 & 100.0 & 100.0 & 100.0 & 99.8 & 100.0 & 99.6 & 100.0 & 100.0 \\
2 & 493 & 100.0 & 100.0 & 100.0 & 99.8 & 100.0 & 96.3 & 100.0 & 100.0 \\
3 & 483 & 100.0 & 100.0 & 100.0 & 98.6 & 100.0 & 90.7 & 99.8 & 100.0 \\
4 & 419 & 100.0 & 100.0 & 100.0 & 98.6 & 98.8 & 89.5 & 100.0 & 100.0 \\
5 & 396 & 100.0 & 100.0 & 100.0 & 97.2 & 97.0 & 84.6 & 96.5 & 98.2 \\
6 & 325 & 100.0 & 100.0 & 100.0 & 96.0 & 93.8 & 61.2 & 89.2 & 96.9 \\
7 & 266 & 100.0 & 100.0 & 100.0 & 96.2 & 86.5 & 53.0 & 82.3 & 94.4 \\
8 & 259 & 100.0 & 100.0 & 99.6 & 91.9 & 77.2 & 34.0 & 66.4 & 84.6 \\
9 & 240 & 100.0 & 100.0 & 100.0 & 90.0 & 65.8 & 17.9 & 56.7 & 72.9 \\
10 & 194 & 100.0 & 99.5 & 95.4 & 90.7 & 51.5 & 12.9 & 36.6 & 58.2 \\
11 & 182 & 100.0 & 100.0 & 95.1 & 78.0 & 28.0 & 13.2 & 29.7 & 48.9 \\
12 & 127 & 100.0 & 100.0 & 89.8 & 75.6 & 21.3 & 17.3 & 33.9 & 40.9 \\
13 & 91 & 100.0 & 98.9 & 90.1 & 51.6 & 6.6 & 4.4 & 18.7 & 20.9 \\
14 & 55 & 98.2 & 98.2 & 78.2 & 40.0 & 0.0 & 9.1 & 16.4 & 20.0 \\
15 & 4096 & 98.8 & 98.5 & 74.9 & 42.9 & 0.6 & 4.8 & 19.9 & 24.7 \\
\bottomrule
\end{tabular}
\caption{In-distribution answer accuracy (\%) per number of operations for the clean model, the shuffled-prefix models ($p=10$ to $100$), the swapped model, and the model trained with no trace.}
\label{tab:arms-shuffled-per-op}
\end{table}

\begin{table}[h]
\centering
\small
\setlength{\tabcolsep}{3.5pt}
\begin{tabular}{l rr rr rr rr rr}
\toprule
 & \multicolumn{2}{c}{op $=15$} & \multicolumn{2}{c}{op $=20$} & \multicolumn{2}{c}{op $=21$} & \multicolumn{2}{c}{op $=22$} & \multicolumn{2}{c}{op $=23$} \\
\cmidrule(lr){2-3}\cmidrule(lr){4-5}\cmidrule(lr){6-7}\cmidrule(lr){8-9}\cmidrule(lr){10-11}
 & correct & wrong & correct & wrong & correct & wrong & correct & wrong & correct & wrong \\
\midrule
trace valid   & 4027 &   20 & 3154 &  132 & 2788 &  115 & 2181 &  141 & 1626 &  109 \\
trace invalid &   20 &   29 &  256 &  554 &  397 &  796 &  595 & 1179 &  750 & 1611 \\
\bottomrule
\end{tabular}
\caption{For the clean model, a quadrant of trace validity versus answer correctness for all the operations. We find that at larger operations, the correlation between trace validity and answer correctness drops.}
\label{tab:quadrants-clean-full}
\end{table}

\begin{table}[h]
\centering
\footnotesize
\setlength{\tabcolsep}{4pt}
\begin{tabular}{l r r r r r l}
\toprule
 & op $=15$ & op $=20$ & op $=21$ & op $=22$ & op $=23$ & example \\
\midrule
correct answers & 4047 & 3410 & 3185 & 2776 & 2376 &  \\
\quad with an invalid trace & 20 & 256 & 397 & 595 & 750 &  \\
\quad share of correct answers (\%) & 0.5 & 7.5 & 12.5 & 21.4 & 31.6 &  \\
\midrule
\textbf{Syntactic/arithmetic failures (total)} & 1 & 56 & 146 & 267 & 366 &  \\
\quad share of invalid traces (\%) & 5.0 & 21.9 & 36.8 & 44.9 & 48.8 &  \\
\quad trace does not parse & 1 & 54 & 141 & 259 & 352 & Ex.~3 \\
\quad reference, order or redefinition error & 0 & 2 & 5 & 8 & 14 & Ex.~4 \\
\quad arithmetic error & 0 & 0 & 0 & 0 & 0 &  \\
\midrule
\textbf{Semantic-only failures (total)} & 19 & 200 & 251 & 328 & 384 &  \\
\quad share of invalid traces (\%) & 95.0 & 78.1 & 63.2 & 55.1 & 51.2 &  \\
\quad defines a name not in the problem & 0 & 3 & 5 & 6 & 6 & Ex.~5 \\
\quad a step uses the wrong inputs & 19 & 197 & 246 & 320 & 377 &  \\
\qquad same-value substitution or result & 14 & 130 & 160 & 214 & 234 & Ex.~2 \\
\qquad wrong value later multiplied by zero & 4 & 61 & 74 & 90 & 119 & Ex.~1 \\
\qquad wrong value reaches the answer, matches & 1 & 5 & 11 & 13 & 20 & Ex.~6 \\
\qquad wrong value never used again & 0 & 1 & 1 & 3 & 4 & Ex.~7 \\
\quad other dependency mismatch & 0 & 0 & 0 & 2 & 1 & Ex.~8 \\
\bottomrule
\end{tabular}
\caption{Table~\ref{tab:invalid-correct-causes} extended to all five operation levels. The row ``share of correct answers'' is the number of invalid traces divided by the number of correct answers. The two ``share of invalid traces'' rows divide each group's total by the number of invalid traces, the row ``with an invalid trace''. The last column points to an example of each row in Appendix~\ref{sec:examples}.}
\label{tab:invalid-correct-causes-full}
\end{table}

\begin{table}[h]
\centering
\footnotesize
\setlength{\tabcolsep}{4pt}
\begin{tabular}{l r l rrrrr}
\toprule
clean model & steps & setup & op $=15$ & op $=20$ & op $=21$ & op $=22$ & op $=23$ \\
\midrule
Ye et al.\ (reported) & 100k & trace valid too & 99.1 & 91.8 & 87.9 & 84.0 & 76.8 \\
\midrule
run A                 & 100k & used in all tables    & 98.8 & 83.3 & 77.8 & 67.8 & 58.0 \\
                      &      & trace valid too       & 98.3 & 77.0 & 68.1 & 53.2 & 39.7 \\
run B                 & 100k & second run            & 99.1 & 81.3 & 71.7 & 57.2 & 47.3 \\
run C                 & 200k & twice the steps       & 99.7 & 87.4 & 80.4 & 68.5 & 58.5 \\
                      &      & trace valid too       & 99.6 & 83.9 & 73.6 & 57.3 & 41.3 \\
\bottomrule
\end{tabular}
\caption{Accuracy (\%) of our clean models against the values reported by \citet{ye2024physics21} for iGSM-med; all runs use their recipe. Ye et al.\ count a solution only when its trace also passes their checker; the rows marked ``trace valid too'' apply the same criterion to our models. Runs A and B differ by ten points at op $=23$.}
\label{tab:clean-runs}
\end{table}

\begin{table}[h]
\centering
\begin{tabular}{l r rr rrrr}
\toprule
 & & \multicolumn{2}{c}{in-distribution} & \multicolumn{4}{c}{out of distribution} \\
\cmidrule(lr){3-4}\cmidrule(lr){5-8}
training trace & seed & op $\le 15$ & op $=15$ & op $=20$ & op $=21$ & op $=22$ & op $=23$ \\
\midrule
swapped              & 42 & 81.7 & 19.9 & 16.5 & 16.9 & 16.8 & 17.4 \\
                     & 43 & 79.9 & 19.1 & 16.4 & 16.3 & 16.6 & 17.2 \\
swapped, op-matched  & 42 & 81.4 & 19.2 & 16.1 & 16.4 & 16.7 & 17.1 \\
                     & 43 & 81.2 & 18.3 & 16.6 & 16.3 & 16.6 & 16.7 \\
30\% shuffled        & 42 & 98.6 & 74.9 & 43.5 & 37.3 & 31.2 & 27.8 \\
                     & 43 & 96.7 & 69.4 & 38.1 & 34.0 & 28.6 & 24.2 \\
100\% non-minimal    & 42 & 98.9 & 86.5 & 75.3 & 72.6 & 69.8 & 68.7 \\
                     & 43 & 98.3 & 90.1 & 78.2 & 76.3 & 70.6 & 67.0 \\
\bottomrule
\end{tabular}
\caption{Answer accuracy (\%) of a second training seed for the models trained on swapped, shuffled and non-minimal traces. The seed-42 rows are the models reported in the main text; the seed-43 rows use the same recipe. Accuracies agree within five points at every level and the ordering of the models is unchanged.}
\label{tab:seed-replicates}
\end{table}

\section{Example traces}
\label{sec:examples}
Examples 1 to 9 are from the clean model. Examples 1 to 8 show correct answers with invalid traces, one for each row of Table~\ref{tab:invalid-correct-causes-full}; the table's last column says which. Example 9 is a valid trace with a wrong answer. Examples 10 to 14 are from the swapped, shuffled and non-minimal models, and Example 15 is the clean model on a re-asked question.

\paragraph{Example 1. Clean model, correct answer with an invalid trace. Semantic failure: one step uses the wrong input, and the error is then multiplied by zero.}
\begin{quote}\small
\textbf{Problem.} The number of each Pericardium's Monocytes equals the sum of each Cheetah's Organs and each Cheetah's Mitral Valve. The number of each Cheetah's Mitral Valve equals each Great Victoria Desert's Cells. The number of each Gobi Desert's Cheetah equals 11 times as much as each Great Victoria Desert's Creatures. The number of each Gorilla's Pericardium equals the sum of each Gobi Desert's Creatures, each Gobi Desert's Gorilla and each Gobi Desert's Cheetah. The number of each Pericardium's Eosinophils equals the sum of each Gorilla's Organs, each Gorilla's Pericardium and each Pericardium's Monocytes. The number of each Gobi Desert's Gorilla equals 12. How many Cells does Gobi Desert have?

\textbf{Model trace.} Define Great Victoria Desert's Creatures as a; so a = 0. Define Gobi Desert's Cheetah as b; so b = 11 * a = 11 * 0 = 0. Define Gobi Desert's Gorilla as P; so P = 12. Define Gobi Desert's Creatures as p; so p = b + P = 0 + 12 = 12. Define Gorilla's Pericardium as o; B = p + P = 12 + 12 = 1; so o = B + b = 1 + 0 = 1. Define Gorilla's Organs as c; so c = o = 1. Define Great Victoria Desert's Cells as w; so w = 0. Define Cheetah's Mitral Valve as m; so m = w = 0. Define Cheetah's Organs as f; so f = m = 0. Define Pericardium's Monocytes as r; so r = f + m = 0 + 0 = 0. Define Pericardium's Eosinophils as K; A = c + o = 1 + 1 = 2; so K = A + r = 2 + 0 = 2. Define Pericardium's Cells as X; so X = r + K = 0 + 2 = 2. \textbf{Define Cheetah's Cells as k; so k = m * X = 0 * 2 = 0.} Define Gorilla's Cells as S; so S = o * X = 1 * 2 = 2. Define Gobi Desert's Cells as H; M = b * k = 0 * 0 = 0; j = P * S = 12 * 2 = 1; so H = M + j = 0 + 1 = 1.

\textbf{Answer.} 1

\textbf{Checker.} Invalid, at the step in bold. Cheetah's only organ is Mitral Valve, so Cheetah's Cells should be Cheetah's Mitral Valve (m) times Mitral Valve's Cells. The trace instead multiplies m by X, which is Pericardium's Cells, the total for a different organ. Because m = 0, the product is 0 either way, and the answer is unaffected.
\end{quote}

\paragraph{Example 2. Clean model, correct answer with an invalid trace. Semantic failure: one step uses the wrong input, which happens to have the same value as the right one.}
\begin{quote}\small
\textbf{Problem.} The number of each Arctic Ocean's Stork equals 3 times as much as the sum of each Tidal Pool's Creatures and each Owl's Autonomic Nerves. The number of each Arctic Ocean's Owl equals 0. The number of each Autonomic Nerves's Plasma Cells equals 22 more than the sum of each Arctic Ocean's Creatures and each Arctic Ocean's Owl. The number of each Autonomic Nerves's Basophils equals the difference of each Tidal Pool's Organs and each Tidal Pool's Creatures. \uline{The number of each Stork's Thalamus equals 7 more than each Owl's Organs.} The number of each Owl's Autonomic Nerves equals the sum of each Tidal Pool's Cells and each Arctic Ocean's Owl. How many Cells does Arctic Ocean have?

\textbf{Model trace.} Define Arctic Ocean's Owl as f; so f = 0. Define Tidal Pool's Cells as s; so s = 0. Define Owl's Autonomic Nerves as R; so R = s + f = 0 + 0 = 0. Define Tidal Pool's Creatures as W; so W = 0. Define Arctic Ocean's Stork as X; v = W + R = 0 + 0 = 0; so X = 3 * v = 3 * 0 = 0. Define Arctic Ocean's Creatures as r; so r = f + X = 0 + 0 = 0. Define Autonomic Nerves's Plasma Cells as O; P = r + f = 0 + 0 = 0; so O = 22 + P = 22 + 0 = 22. Define Tidal Pool's Organs as J; so J = 0. Define Autonomic Nerves's Basophils as x; so x = J - W = 0 - 0 = 0. Define Autonomic Nerves's Cells as A; so A = O + x = 22 + 0 = 22. Define Owl's Cells as U; so U = R * A = 0 * 22 = 0. \textbf{Define Stork's Thalamus as o; so o = 7 + U = 7 + 0 = 7.} Define Thalamus's Cells as k; so k = 0. Define Stork's Cells as L; so L = o * k = 7 * 0 = 0. Define Arctic Ocean's Cells as b; q = f * U = 0 * 0 = 0; z = X * L = 0 * 0 = 0; so b = q + z = 0 + 0 = 0.

\textbf{Answer.} 0

\textbf{Checker.} Invalid, at the step in bold. The underlined sentence says Stork's Thalamus is 7 more than Owl's Organs, but the trace adds 7 to U, which is Owl's Cells. Both are 0, so every later value, and the answer, is unchanged.
\end{quote}

\paragraph{Example 3. Clean model, correct answer with an invalid trace. Syntactic failure: the trace does not parse, because it uses a symbol it never defined.}
\begin{quote}\small
\textbf{Problem.} The number of each Ecology Lab's Sweat-Resistant Backpack equals each Harbortown University's Stationery. The number of each Eastwood University's Botany Lab equals 7 times as much as each Ecology Lab's Backpack. The number of each Golf Backpack's Correction Tape equals each Eastwood University's Classroom. The number of each Eastwood University's Ecology Lab equals 2 times as much as each Ecology Lab's Sweat-Resistant Backpack. The number of each Ecology Lab's Golf Backpack equals 8 times as much as the difference of each Harbortown University's Stationery and each Ecology Lab's Sweat-Resistant Backpack. The number of each Sweat-Resistant Backpack's Binder Clip equals each Harbortown University's Backpack. How many Stationery does Eastwood University have?

\textbf{Model trace.} Define Harbortown University's Stationery as u; so u = 0. Define Ecology Lab's Sweat-Resistant Backpack as g; so g = u = 0. Define Ecology Lab's Golf Backpack as R; W = u - g = 0 - 0 = 0; so R = 8 * W = 8 * 0 = 0. Define Ecology Lab's Backpack as o; so o = R + g = 0 + 0 = 0. Define Eastwood University's Botany Lab as z; so z = 7 * o = 7 * 0 = 0. Define Eastwood University's Ecology Lab as F; so F = 2 * g = 2 * 0 = 0. Define Eastwood University's Classroom as G; so G = F + z = 0 + 0 = 0. Define Golf Backpack's Correction Tape as w; so w = G = 0. Define Golf Backpack's Stationery as m; so m = w = 0. Define Botany Lab's Stationery as L; so L = 0. \textbf{Define Ecology Lab's Stationery as A; T = R * U = 0 * 0 = 0; K = g * m = 0 * 0 = 0; so A = T + K = 0 + 0 = 0.} Define Eastwood University's Stationery as c; v = F * A = 0 * 0 = 0; B = z * L = 0 * 0 = 0; so c = v + B = 0 + 0 = 0.

\textbf{Answer.} 0

\textbf{Checker.} Invalid, at the step in bold. The clause T = R * U uses a capital U that the trace never defined; it defined only lowercase u, with value 0. Every quantity in this problem is 0, so the answer is unaffected.
\end{quote}

\paragraph{Example 4. Clean model, correct answer with an invalid trace. Syntactic failure: a reference error, the trace skips a parameter and names the wrong symbol in its place.}
\begin{quote}\small
\textbf{Problem.} The number of each Tidal Pool's Dragonfly equals each Dragonfly's Laryngopharynx. The number of each Tidal Pool's Woodpecker equals 13 times as much as each Dragonfly's Laryngopharynx. The number of each Arctic Ocean's Dragonfly equals 3 more than each Tidal Pool's Organs. The number of each Tidal Pool's Pigeon equals each Coastal Ecosystem's Organs. The number of each Woodpecker's Pleural Cavity equals each Coastal Ecosystem's Organs. The number of each Arctic Ocean's Pigeon equals the difference of each Tidal Pool's Dragonfly and each Dragonfly's Laryngopharynx. The number of each Dragonfly's Laryngopharynx equals 17. The number of each Woodpecker's Oropharynx equals 4 times as much as each Coastal Ecosystem's Creatures. How many Organs does Arctic Ocean have?

\textbf{Model trace.} Define Dragonfly's Laryngopharynx as a; so a = 17. Define Tidal Pool's Dragonfly as p; so p = a = 17. Define Coastal Ecosystem's Organs as b; so b = 0. Define Dragonfly's Organs as o; so o = a = 17. Define Tidal Pool's Pigeon as r; so r = b = 0. Define Pigeon's Organs as B; so B = 0. Define Tidal Pool's Woodpecker as f; so f = 13 * a = 13 * 17 = 14. \textbf{Define Tidal Pool's Organs as x; W = p * o = 17 * 17 = 13; h = r * B = 0 * 0 = 0; q = W + h = 13 + 0 = 13; Y = f * q = 14 * 0 = 0; so x = q + Y = 13 + 0 = 13.} Define Arctic Ocean's Dragonfly as S; so S = 3 + x = 3 + 13 = 16. Define Arctic Ocean's Pigeon as A; so A = p - a = 17 - 17 = 0. Define Arctic Ocean's Organs as G; K = S * o = 16 * 17 = 19; J = A * B = 0 * 0 = 0; so G = K + J = 19 + 0 = 19.

\textbf{Answer.} 19

\textbf{Checker.} Invalid, at the step in bold. The clause should multiply Tidal Pool's Woodpecker, f, by Woodpecker's Organs, which is 0 and which the trace never defines. Instead it writes f * q, naming the partial sum q just computed, whose value is 13, while substituting the 0 it needed. The checker flags the reference to q. Because the intended factor is 0 either way, the product and the answer are unaffected.
\end{quote}

\paragraph{Example 5. Clean model, correct answer with an invalid trace. Semantic failure: the trace defines a name not in the problem.}
\begin{quote}\small
\textbf{Problem.} The number of each Historic Industrial District's Boxed equals each Historic Downtown's DoorDash. The number of each DoorDash's Sparkling Water equals each Wine's Cherry. The number of each Deliveroo's Wine equals 13 times as much as the sum of each Historic Industrial District's Supermarket and each Historic Downtown's DoorDash. The number of each Wine's Cherry equals 1. The number of each Wine's Mango equals the sum of each Quarters's Product, each Quarters's Deliveroo, each Deliveroo's Product and each Wine's Cherry. The number of each Sparkling Water's Lemon equals 13. The number of each Deliveroo's Almond Milk equals 11 more than the difference of each DoorDash's Wine and each Almond Milk's Ingredient. The number of each Historic Downtown's DoorDash equals 16 times as much as the difference of each DoorDash's Sparkling Water and each Wine's Cherry. The number of each DoorDash's Wine equals 14 times as much as the difference of each Historic Industrial District's Boxed and each DoorDash's Sparkling Water. \uline{The number of each Quarters's Deliveroo equals 7 more than the difference of each Almond Milk's Ingredient and each Historic Downtown's DoorDash.} The number of each Historic Downtown's Boxed equals the sum of each Historic Downtown's DoorDash, each DoorDash's Ingredient, each Wine's Cherry and each Quarters's Ingredient. How many Mango does Wine have?

\textbf{Model trace.} Define Wine's Cherry as c; so c = 1. Define DoorDash's Sparkling Water as w; so w = c = 1. Define Historic Downtown's DoorDash as A; B = w - c = 1 - 1 = 0; so A = 16 * B = 16 * 0 = 0. Define Historic Industrial District's Boxed as K; so K = A = 0. Define DoorDash's Wine as b; J = K - w = 0 - 1 = 22; so b = 14 * J = 14 * 22 = 9. Define Almond Milk's Ingredient as M; so M = 0. Define Deliveroo's Almond Milk as G; f = b - M = 9 - 0 = 9; so G = 11 + f = 11 + 9 = 20. \textbf{Define Historic Downtown's Deliveroo as P; z = M - A = 0 - 0 = 0; so P = 7 * z = 7 * 0 = 0.} Define Quarters's Deliveroo as r; h = M - A = 0 - 0 = 0; so r = 7 + h = 7 + 0 = 7. Define Historic Industrial District's Supermarket as E; so E = K = 0. Define Deliveroo's Wine as a; O = E + A = 0 + 0 = 0; so a = 13 * O = 13 * 0 = 0. Define Deliveroo's Product as X; so X = G + a = 20 + 0 = 20. Define Quarters's Product as R; so R = r * X = 7 * 20 = 2. Define Wine's Mango as k; S = R + r = 2 + 7 = 9; q = S + X = 9 + 20 = 6; so k = q + c = 6 + 1 = 7.

\textbf{Answer.} 7

\textbf{Checker.} Invalid, at the step in bold. No sentence of the problem mentions Historic Downtown's Deliveroo. The trace defines it anyway, from the inputs of the underlined sentence, Almond Milk's Ingredient and Historic Downtown's DoorDash, but with 7 times their difference where the sentence says 7 more than it. The checker rejects the unknown name. The next step defines Quarters's Deliveroo correctly, the invented symbol P is never read, and the answer is unaffected.
\end{quote}

\paragraph{Example 6. Clean model, correct answer with an invalid trace. Semantic failure: one step uses the wrong input, and the error reaches the answer but cancels.}
\begin{quote}\small
\textbf{Problem.} The number of each Amboseli National Park's Zebra Paddock equals 7. \uline{The number of each Serengeti National Park's Cheetah Run equals 9 times as much as the difference of each Serengeti National Park's Zebra Paddock and each Amboseli National Park's Animal.} The number of each Bat Cave's Squirrel equals 9 times as much as the sum of each Zebra Paddock's Squirrel and each Zebra Paddock's Animal. The number of each Amboseli National Park's Bat Cave equals 13 more than the sum of each Zebra Paddock's Animal and each Otter Pool's Kangaroo. The number of each Zebra Paddock's Squirrel equals 7 more than the difference of each Amboseli National Park's Otter Pool and each Otter Pool's Kangaroo. \uline{The number of each Serengeti National Park's Zebra Paddock equals 5 more than each Amboseli National Park's Animal.} The number of each Amboseli National Park's Otter Pool equals each Otter Pool's Kangaroo. The number of each Otter Pool's Kangaroo equals 22. How many Cheetah Run does Serengeti National Park have?

\textbf{Model trace.} Define Otter Pool's Kangaroo as a; so a = 22. Define Amboseli National Park's Otter Pool as b; so b = a = 22. Define Zebra Paddock's Squirrel as P; z = b - a = 22 - 22 = 0; so P = 7 + z = 7 + 0 = 7. Define Zebra Paddock's Animal as p; so p = P = 7. Define Amboseli National Park's Bat Cave as G; Y = p + a = 7 + 22 = 6; so G = 13 + Y = 13 + 6 = 19. Define Bat Cave's Squirrel as S; h = P + p = 7 + 7 = 14; so S = 9 * h = 9 * 14 = 11. Define Bat Cave's Animal as k; so k = S = 11. Define Amboseli National Park's Zebra Paddock as r; so r = 7. \textbf{Define Amboseli National Park's Animal as g; q = G * p = 19 * 7 = 18; B = b * k = 22 * 11 = 12; c = q + B = 18 + 12 = 7; s = r * k = 7 * 11 = 8; so g = c + s = 7 + 8 = 15.} Define Serengeti National Park's Zebra Paddock as J; so J = 5 + g = 5 + 15 = 20. Define Serengeti National Park's Cheetah Run as K; f = J - g = 20 - 15 = 5; so K = 9 * f = 9 * 5 = 22.

\textbf{Answer.} 22

\textbf{Checker.} Invalid, at the step in bold. Amboseli National Park's Animal should be the sum over its three exhibits of the exhibit count times that exhibit's Animal, $19\times11 + 7\times7 + 22\times22$, which is 6. The trace pairs every exhibit with another exhibit's Animal and never reads Otter Pool's Animal, so the dependency check fails and the step gives 15. The wrong value flows on through the two underlined sentences. Serengeti National Park's Zebra Paddock is 5 more than it, and Cheetah Run is 9 times the difference of the two. That difference is 5 whatever the value, so the answer is 22 either way.
\end{quote}

\paragraph{Example 7. Clean model, correct answer with an invalid trace. Semantic failure: one step uses the wrong input, and the value is never used again.}
\begin{quote}\small
\textbf{Problem.} The number of each Eastside Secondary's Public Administration Classroom equals 20 more than the sum of each Pinnacle High's Stationery, each Sociology Lab's Casual Backpack and each Sociology Lab's Backpack. The number of each Public Administration Classroom's Athletic Backpack equals 2 times as much as the difference of each Pinnacle High's Backpack and each Pinnacle High's Stationery. \uline{The number of each Athletic Backpack's Document Box equals each Pinnacle High's Classroom.} The number of each Sociology Lab's Casual Backpack equals 15. The number of each Eastside Secondary's Sociology Lab equals 2 more than the sum of each Sociology Lab's Backpack and each Pinnacle High's Stationery. The number of each Athletic Backpack's Magazine Holder equals each Eastside Secondary's Classroom. How many Stationery does Eastside Secondary have?

\textbf{Model trace.} Define Sociology Lab's Casual Backpack as u; so u = 15. Define Sociology Lab's Backpack as z; so z = u = 15. Define Pinnacle High's Stationery as R; so R = 0. Define Eastside Secondary's Public Administration Classroom as J; s = R + u = 0 + 15 = 15; G = s + z = 15 + 15 = 7; so J = 20 + G = 20 + 7 = 4. Define Pinnacle High's Backpack as S; so S = 0. Define Public Administration Classroom's Athletic Backpack as g; F = S - R = 0 - 0 = 0; so g = 2 * F = 2 * 0 = 0. Define Eastside Secondary's Sociology Lab as f; p = z + R = 15 + 0 = 15; so f = 2 + p = 2 + 15 = 17. \textbf{Define Athletic Backpack's Document Box as W; so W = f = 17.} Define Casual Backpack's Stationery as o; so o = 0. Define Public Administration Classroom's Stationery as c; so c = g * o = 0 * 0 = 0. Define Sociology Lab's Stationery as A; so A = u * o = 15 * 0 = 0. Define Eastside Secondary's Stationery as X; L = f * A = 17 * 0 = 0; v = J * c = 4 * 0 = 0; so X = L + v = 0 + 0 = 0.

\textbf{Answer.} 0

\textbf{Checker.} Invalid, at the step in bold. The underlined sentence sets Athletic Backpack's Document Box equal to Pinnacle High's Classroom, which is 0. The trace instead copies the value of the parameter defined just before, Eastside Secondary's Sociology Lab, 17, and the dependency check fails. The symbol W is never read again, because the trace skips Athletic Backpack's Stationery, which would have used it. It computes Public Administration Classroom's Stationery from Casual Backpack's Stationery instead, a second wrong input. There are 0 athletic backpacks, so that product is 0 either way, and the answer is unaffected.
\end{quote}

\paragraph{Example 8. Clean model, correct answer with an invalid trace. Semantic failure, other dependency mismatch: one clause reads an intermediate result of an earlier step instead of a parameter.}
\begin{quote}\small
\textbf{Problem.} The number of each Georgia Aquarium's Aviary equals the sum of each National Aquarium's Animal, each Penguin Exhibit's Salamander, each National Aquarium's Aviary and each National Aquarium's Macaw Island. The number of each National Aquarium's Aviary equals the sum of each Aviary's Animal, each Penguin Exhibit's Salamander, each Penguin Exhibit's Animal and each Penguin Exhibit's Tree Frog. The number of each Penguin Exhibit's Salamander equals 8 times as much as the difference of each Penguin Exhibit's Tree Frog and each National Aquarium's Macaw Island. The number of each National Aquarium's Macaw Island equals 22 times as much as each Penguin Exhibit's Tree Frog. The number of each Penguin Exhibit's Tree Frog equals 0. The number of each National Aquarium's Penguin Exhibit equals 3 more than the sum of each Penguin Exhibit's Tree Frog, each National Aquarium's Macaw Island and each Penguin Exhibit's Salamander. How many Animal does Georgia Aquarium have?

\textbf{Model trace.} Define Penguin Exhibit's Tree Frog as c; so c = 0. Define National Aquarium's Macaw Island as w; so w = 22 * c = 22 * 0 = 0. Define Penguin Exhibit's Salamander as A; B = c - w = 0 - 0 = 0; so A = 8 * B = 8 * 0 = 0. Define National Aquarium's Penguin Exhibit as K; M = c + w = 0 + 0 = 0; z = M + A = 0 + 0 = 0; so K = 3 + z = 3 + 0 = 3. Define Penguin Exhibit's Animal as G; so G = c + A = 0 + 0 = 0. Define Aviary's Animal as P; so P = 0. Define National Aquarium's Aviary as S; O = P + A = 0 + 0 = 0; I = O + G = 0 + 0 = 0; so S = I + c = 0 + 0 = 0. \textbf{Define National Aquarium's Animal as q; J = S * P = 0 * 0 = 0; n = K * G = 3 * 0 = 0; f = J + n = 0 + 0 = 0; h = w * O = 0 * 0 = 0; so q = f + h = 0 + 0 = 0.} Define Georgia Aquarium's Aviary as a; b = q + A = 0 + 0 = 0; X = b + S = 0 + 0 = 0; so a = X + w = 0 + 0 = 0. Define Georgia Aquarium's Animal as R; so R = a * P = 0 * 0 = 0.

\textbf{Answer.} 0

\textbf{Checker.} Invalid, at the step in bold. National Aquarium's Animal should be the sum over its three exhibits of the exhibit count times that exhibit's Animal. The third product, h = w * O, multiplies National Aquarium's Macaw Island by O instead of by Macaw Island's Animal, which the trace never defines. O is not a parameter but a scratch symbol, the running sum inside the earlier step for National Aquarium's Aviary. Because the wrong input is not a parameter of the problem, the table counts the case as ``other'' rather than as a wrong input. Every quantity here is 0, so the answer is unaffected.
\end{quote}

\paragraph{Example 9. Clean model, wrong answer with a valid trace. The last step reads the right parameters but pairs them in the wrong products.}
\begin{quote}\small
\textbf{Problem.} The number of each Lizard's Caudal Vertebrae 8 equals 0 more than each Lizard's Caudal Vertebrae 7. The number of each Gecko's Caudal Vertebrae 1 equals 13 times as much as each Lizard's Caudal Vertebrae 7. The number of each Anteater Area's Lizard equals 7 times as much as the sum of each Gecko's Caudal Vertebrae 1 and each Lizard's Caudal Vertebrae 7. The number of each Anteater Area's Gecko equals 2. The number of each Anteater Area's Crocodile equals 7. The number of each Lizard's Caudal Vertebrae 7 equals 4. How many Bone does Anteater Area have?

\textbf{Model trace.} Define Lizard's Caudal Vertebrae 7 as f; so f = 4. Define Lizard's Caudal Vertebrae 8 as w; so w = 0 + f = 0 + 4 = 4. Define Gecko's Caudal Vertebrae 1 as X; so X = 13 * f = 13 * 4 = 6. Define Anteater Area's Lizard as g; A = X + f = 6 + 4 = 10; so g = 7 * A = 7 * 10 = 1. Define Lizard's Bone as R; so R = f + w = 4 + 4 = 8. Define Anteater Area's Gecko as T; so T = 2. Define Gecko's Bone as v; so v = X = 6. Define Anteater Area's Crocodile as K; so K = 7. Define Crocodile's Bone as G; so G = 0. \textbf{Define Anteater Area's Bone as j; k = T * v = 2 * 6 = 12; c = K * R = 7 * 8 = 10; N = k + c = 12 + 10 = 22; h = g * G = 1 * 0 = 0; so j = N + h = 22 + 0 = 22.}

\textbf{Answer.} 22 (true answer 20)

\textbf{Checker.} Valid. Anteater Area's Bone is the sum, over its three animals, of the number of that animal times that animal's Bone: T times v, K times G, and g times R. The step in bold reads exactly these six parameters, so the dependency check passes, and every equation it writes is arithmetically right, so the arithmetic check passes. But it multiplies K by R and g by G, pairing the crocodiles with the lizard's bones and the lizards with the crocodile's. The correct value is $2\times6 + 7\times0 + 1\times8 = 20$. All 20 valid traces with wrong answers at $\mathsf{op}=15$, and 124 of the 132 at $\mathsf{op}=20$, are of this kind (Table~\ref{tab:quadrants-clean-full}).
\end{quote}

\paragraph{Example 10. Swapped model, correct answer. The trace belongs to no problem.}
\begin{quote}\small
\textbf{Problem.} The number of each Bass's Metacarpal II equals 11. The number of each Shark's Metacarpal I equals each Bass's Metacarpal II. How many Bone does Shark have?

\textbf{Model trace.} Define Canned Peas's Ingredient as b; so b = 0. Define Canned Broth's Chicken Wing as p; so p = b = 0. Define Canned Broth's Ingredient as m; so m = p = 0.

\textbf{Answer.} 11

\textbf{Checker.} Invalid. The trace names parameters that do not occur in the problem.
\end{quote}

\paragraph{Example 11. Swapped model, a second problem with a correct answer.}
\begin{quote}\small
\textbf{Problem.} The number of each Newt Nook's Gila Monster equals 13. The number of each Newt Nook's Skink equals 4 more than each Newt Nook's Gila Monster. How many Animal does Newt Nook have?

\textbf{Model trace.} Define Canned Peas's Ingredient as b; so b = 0. Define Canned Broth's Chicken Wing as J; so J = b = 0. Define Canned Broth's Ingredient as p; so p = J = 0.

\textbf{Answer.} 7

\textbf{Checker.} Invalid, as above.
\end{quote}

\paragraph{Example 12. Shuffled-prefix model ($p=30$), correct answer.}
\begin{quote}\small
\textbf{Problem.} The number of each Silicon Valley's Mitsuwa Marketplace equals each Wall Street's Supermarket. The number of each Wall Street's Valli Produce equals each Business District's NetCost Market. The number of each Financial District's Buford Highway Farmers Market equals each Business District's Buford Highway Farmers Market. The number of each Silicon Valley's Buford Highway Farmers Market equals 4 times as much as each Silicon Valley's Mitsuwa Marketplace. The number of each Financial District's NetCost Market equals 1 more than each Business District's NetCost Market. The number of each Business District's Buford Highway Farmers Market equals 7 times as much as each Silicon Valley's Supermarket. The number of each Business District's Valli Produce equals 7 times as much as each Financial District's Valli Produce. The number of each Financial District's Valli Produce equals 18 times as much as each Business District's Buford Highway Farmers Market. The number of each Financial District's Mitsuwa Marketplace equals 17. The number of each Wall Street's Buford Highway Farmers Market equals 15 more than each Wall Street's Valli Produce. The number of each Business District's NetCost Market equals 22. The number of each Business District's Mitsuwa Marketplace equals each Silicon Valley's Buford Highway Farmers Market. How many Supermarket does Financial District have?

\textbf{Model trace.} = = = = = Street.'sine so Def; as + = 22.ine so;'s as Def + Street Wall = 22;. so'sine as + 14 Wall Market Def V 13 22 = 15 B Net Street Business ProdualliceCost so; Define 14 Farmers 15 Super as +'s. V V V Districtmarket Marketorduf Highway Wall B B B B V 22 = B V K K K \textbf{Define Silicon Valley's Mitsuwa Marketplace as T; so T = B = 13.} Define Silicon Valley's Buford Highway Farmers Market as v; so v = 4 * T = 4 * 13 = 6. Define Silicon Valley's Supermarket as D; so D = T + v = 13 + 6 = 19. Define Business District's Buford Highway Farmers Market as f; so f = 7 * D = 7 * 19 = 18. Define Financial District's Buford Highway Farmers Market as a; so a = f = 18. Define Financial District's Mitsuwa Marketplace as w; so w = 17. \textbf{Define Financial District's NetCost Market as S; so S = 1 + K = 1 + 22 = 0.} Define Financial District's Valli Produce as F; so F = 18 * f = 18 * 18 = 2. Define Financial District's Supermarket as p; h = w + a = 17 + 18 = 12; r = h + S = 12 + 0 = 12; so p = r + F = 12 + 2 = 14.

\textbf{Answer.} 14

\textbf{Checker.} Invalid. The trace begins with a block of permuted tokens that does not parse; the sentences after it are well formed and every step is arithmetically right. The two steps in bold take B = 13 and K = 22 out of the block; these are the correct values of Business District's Mitsuwa Marketplace and Business District's NetCost Market.
\end{quote}

\paragraph{Example 13. Shuffled-prefix model ($p=30$), wrong answer.}
\begin{quote}\small
\textbf{Problem.} The number of each Arts Campus's Seafood City Supermarket equals 14. The number of each T\&T Supermarket's Swiss Cheese equals each Technical College District's Supermarket. \uline{The number of each La Michoacana Meat Market's Provolone Cheese equals the sum of each Seafood City Supermarket's Parmesan Cheese and each Arts Campus's T\&T Supermarket.} The number of each Liberal Arts College District's La Michoacana Meat Market equals each Special Education District's Product. The number of each Technical College District's Superior Grocers equals each La Michoacana Meat Market's Product. The number of each Seafood City Supermarket's Parmesan Cheese equals the sum of each Arts Campus's Supermarket and each Arts Campus's T\&T Supermarket. The number of each La Michoacana Meat Market's Monterey Jack Cheese equals each Special Education District's Supermarket. The number of each Technical College District's T\&T Supermarket equals 16 times as much as each Liberal Arts College District's Product. The number of each Arts Campus's T\&T Supermarket equals each Arts Campus's Seafood City Supermarket. How many Swiss Cheese does T\&T Supermarket have?

\textbf{Model trace.} = = = = 14 = 14ine's as +market. Def Super; so = 14; +'s as so Define City Super.market Campus = 5 + Arts; 14 Super so Def's asine.market Campusood Seaf Arts =\& Parm Tesan City Super 14 5 19 + DefT; T T T T asine so.market Campus Arts Cheese's =ood Seaf T l l l l l Z Z Z \textbf{Define La Michoacana Meat Market's Provolone Cheese as v; so v = Z + l = 5 + 14 = 19.} Define Special Education District's Supermarket as p; so p = 0. Define La Michoacana Meat Market's Monterey Jack Cheese as a; so a = p = 0. Define La Michoacana Meat Market's Product as M; so M = a + v = 0 + 19 = 19. Define Special Education District's Product as e; so e = 0. Define Liberal Arts College District's La Michoacana Meat Market as f; so f = e = 0. Define Liberal Arts College District's Product as i; so i = f * M = 0 * 19 = 0. Define Technical College District's T\&T Supermarket as g; so g = 16 * i = 16 * 0 = 0. Define Technical College District's Superior Grocers as d; so d = M = 19. Define Technical College District's Supermarket as w; so w = d + g = 19 + 0 = 19. Define T\&T Supermarket's Swiss Cheese as o; so o = w = 19.

\textbf{Answer.} 19 (true answer 10)

\textbf{Checker.} Invalid, and the first well-formed step, in bold, is wrong. The underlined sentence says Provolone Cheese is Parmesan Cheese (19) plus Arts Campus's T\&T Supermarket (14), but the model takes Z = 5, the value of Arts Campus's Supermarket, out of the block. Every later step is arithmetically right and the error reaches the answer.
\end{quote}

\paragraph{Example 14. Model trained on 90\% non-minimal traces, correct answer. The trace is valid but has unnecessary parameters.}
\begin{quote}\small
\textbf{Problem.} The number of each Donuts's Amaranth equals 12. The number of each Food 4 Less's Pies equals the sum of each Shoppers Value Foods's Sourdough Bread and each No Frills's Product. The number of each Sourdough Bread's Quinoa equals 8 times as much as each Donuts's Amaranth. The number of each Quarters's No Frills equals the sum of each Monument Zone's Supermarket and each Food 4 Less's Sourdough Bread. The number of each Shoppers Value Foods's Sourdough Bread equals each No Frills's Donuts. The number of each Food 4 Less's Sourdough Bread equals 4. The number of each Monument Zone's No Frills equals the sum of each Food 4 Less's Sourdough Bread and each No Frills's Donuts. The number of each Old Harbor's Shoppers Value Foods equals the difference of each No Frills's Product and each Donuts's Amaranth. The number of each No Frills's Donuts equals 13. The number of each Quarters's Food 4 Less equals each Old Harbor's Ingredient. The number of each Pies's Buckwheat equals 7 more than each Shoppers Value Foods's Sourdough Bread. How many Supermarket does Quarters have?

\textbf{Model trace.} Define No Frills's Donuts as g; so g = 13. Define Shoppers Value Foods's Sourdough Bread as t; so t = g = 13. \textbf{Define Pies's Buckwheat as h; so h = 7 + t = 7 + 13 = 20.} Define Food 4 Less's Sourdough Bread as E; so E = 4. Define No Frills's Product as y; so y = g = 13. \textbf{Define Food 4 Less's Pies as f; so f = t + y = 13 + 13 = 3.} Define Donuts's Amaranth as z; so z = 12. Define Old Harbor's Shoppers Value Foods as T; so T = y - z = 13 - 12 = 1. Define Monument Zone's No Frills as m; so m = E + g = 4 + 13 = 17. Define Monument Zone's Supermarket as s; so s = m = 17. Define Quarters's No Frills as c; so c = s + E = 17 + 4 = 21. Define Sourdough Bread's Quinoa as e; so e = 8 * z = 8 * 12 = 4. Define Sourdough Bread's Ingredient as k; so k = e = 4. Define Shoppers Value Foods's Ingredient as P; so P = t * k = 13 * 4 = 6. Define Old Harbor's Ingredient as n; so n = T * P = 1 * 6 = 6. Define Quarters's Food 4 Less as O; so O = n = 6. Define Quarters's Supermarket as d; so d = c + O = 21 + 6 = 4.

\textbf{Answer.} 4

\textbf{Checker.} Valid. The two steps in bold define parameters the query does not need, Pies's Buckwheat and Food 4 Less's Pies; every step uses the inputs the problem prescribes.
\end{quote}

\paragraph{Example 15. Clean model, re-asked question, correct answer. The trace is valid but padded with the original question's chain.}
\begin{quote}\small
\textbf{Problem.} The number of each Niagara Falls Aviary's Crustacean Cove equals 17. The number of each Penguin Beach's Sardine equals each Monte Casino Bird Park's Dolphin Lagoon. The number of each Mahi Mahi's Ulna equals the sum of each Penguin Beach's Bone and each Sardine's Metacarpals. The number of each Bird Kingdom's Crustacean Cove equals the sum of each Mahi Mahi's Ulna, each Dolphin Lagoon's Animal, each Penguin Beach's Sardine and each Penguin Beach's Animal. The number of each Niagara Falls Aviary's Penguin Beach equals 17. The number of each Dolphin Lagoon's Sardine equals the difference of each Sardine's Metacarpals and each Monte Casino Bird Park's Dolphin Lagoon. The number of each Penguin Beach's Salmon equals each Sardine's Bone. The number of each Salmon's Coracoid equals 7 more than each Monte Casino Bird Park's Bone. The number of each Monte Casino Bird Park's Dolphin Lagoon equals each Sardine's Metacarpals. The number of each Sardine's Metacarpals equals 20. The number of each Crustacean Cove's Mahi Mahi equals 20. How many Animal does Niagara Falls Aviary have?

\textbf{Model trace.} Define Sardine's Metacarpals as a; so a = 20. Define Monte Casino Bird Park's Dolphin Lagoon as b; so b = a = 20. Define Sardine's Bone as P; so P = a = 20. \textbf{Define Dolphin Lagoon's Sardine as z; so z = a - b = 20 - 20 = 0.} \textbf{Define Dolphin Lagoon's Bone as L; so L = z * P = 0 * 20 = 0.} \textbf{Define Monte Casino Bird Park's Bone as S; so S = b * L = 20 * 0 = 0.} \textbf{Define Salmon's Coracoid as J; so J = 7 + S = 7 + 0 = 7.} Define Penguin Beach's Sardine as p; so p = b = 20. Define Penguin Beach's Salmon as I; so I = P = 20. Define Penguin Beach's Animal as q; so q = p + I = 20 + 20 = 17. Define Niagara Falls Aviary's Penguin Beach as B; so B = 17. Define Crustacean Cove's Mahi Mahi as W; so W = 20. Define Crustacean Cove's Animal as w; so w = W = 20. Define Niagara Falls Aviary's Crustacean Cove as m; so m = 17. Define Niagara Falls Aviary's Animal as c; f = B * q = 17 * 17 = 13; D = m * w = 17 * 20 = 18; so c = f + D = 13 + 18 = 8.

\textbf{Answer.} 8

\textbf{Checker.} Valid. The problem was generated for the question ``Mahi Mahi's Bone'' and re-asked as ``Niagara Falls Aviary's Animal''. The four steps in bold define parameters the new question does not need, all of them on the original question's chain.
\end{quote}

\end{document}